\documentclass{article}
\usepackage{amsmath,amsfonts}
\usepackage{url}
\usepackage{graphicx}
\def\BibTeX{{\rm B\kern-.05em{\sc i\kern-.025em b}\kern-.08em
    T\kern-.1667em\lower.7ex\hbox{E}\kern-.125emX}}
\usepackage{balance}

\usepackage{pdflscape}

\usepackage{IEEEtrantools}
\usepackage{rotating}

\usepackage{times}
\usepackage{latexsym}
\usepackage{mathrsfs}
\usepackage{bbm}
\usepackage{mathbbol}

\usepackage{rotating}
\usepackage{tablefootnote}

\usepackage{url}            
\usepackage{xcolor}
\usepackage{colortbl}
 \usepackage{fancybox}

\usepackage{stackengine}

\usepackage{mathabx}

\usepackage{slashbox}

\usepackage[pagebackref=true,breaklinks=true,colorlinks,bookmarks=false]{hyperref}
\usepackage{tikz}
\usetikzlibrary{positioning,arrows.meta}

\newcommand\xrowht[2][0]{\addstackgap[.5\dimexpr#2\relax]{\vphantom{#1}}}
\newcommand\red[1]{\textcolor{red}{#1}}
\newcommand\blue[1]{\textcolor{blue}{#1}}

\newcommand\magenta[1]{\textcolor{magenta}{#1}}

\definecolor{myblue}{rgb}{0.04,0.23, 1}
\newcommand{\myblue}{\textcolor{myblue}}
\definecolor{mycyan}{rgb}{0.44,0.44,0.2}
\definecolor{myred}{rgb}{0.417         0         0}
\newcommand{\myred}{\textcolor{myred}}

\definecolor{myviolet}{rgb}{0.6     0.4     0.8}

\definecolor{myorange}{rgb}{0.72     0.53     0.08}

\definecolor{mybrown}{rgb}{0.4, 0.26, 0.13}

\definecolor{mygreen}{rgb}{0.6,0.8,0}

\definecolor{mygreenlight}{rgb}{0.25,1,0.50}

\definecolor{redish}{RGB}{255, 88, 0}
\newcommand{\redish}{\textcolor{redish}}

\def\pterror#1{\errmessage{Parsetree ERROR: #1}}
\newdimen\pthgap\def\pthorgap#1{\pthgap=#1}
\newdimen\ptvgap\def\ptvergap#1{\ptvgap=#1}
\newbox\ptnodestrutbox\def\ptnodestrut{\unhcopy\ptnodestrutbox}
\newbox\ptleafstrutbox\def\ptleafstrut{\unhcopy\ptleafstrutbox}
\def\ptnodefont#1#2#3{\def\ptnodefn{#1}
  \setbox\ptnodestrutbox=\hbox{\vrule height#2 width0pt depth#3}}
\def\ptleaffont#1#2#3{\def\ptleaffn{#1}
  \setbox\ptleafstrutbox=\hbox{\vrule height#2 width0pt depth#3}}
\pthorgap{10pt}                         
\ptvergap{28pt}                         
\ptnodefont{\normalsize\rm}{8pt}{3pt}  
\ptleaffont{\normalsize\it}{8pt}{3pt}  
\newcount\ptdepth           
\newcount\ptn               
\newbox\ptm \newdimen\ptmx  
\newbox\pta \newdimen\ptax  
\newbox\ptb \newdimen\ptbx  
\newbox\ptc \newdimen\ptcx  
\newbox\ptx \newdimen\ptxx  
\newif\ifpttri              
\def\ptnext{\advance\ptn by 1 \ifcase\ptn
  \or \setbox\ptm=\box\ptx \ptmx=\ptxx \or \setbox\pta=\box\ptx \ptax=\ptxx
  \or \setbox\ptb=\box\ptx \ptbx=\ptxx \or \setbox\ptc=\box\ptx \ptcx=\ptxx
  \else \pterror{More than 3 daughters in (sub)tree}\fi}
\def\ptbegtree{\ptdepth=0}
\def\ptendtree
  {\ifnum\ptdepth>0 \pterror{Mismatched bracketing: too few ')'s!}\fi}
\def\ptbeg{\ifnum\ptdepth=0 \leavevmode\fi\begingroup
  \advance\ptdepth1 \ptn=0\pttrifalse}
\def\ptend{\ifnum\ptdepth=0 \pterror{Mismatched bracketing: too many ')'s!}
  \else\ptcons\endgroup\ifnum\ptdepth=0 \box\ptx\else\ptnext\fi\fi}
\def\ptnodeaux#1{\setbox\ptx=\hbox{#1}\ptxx=0.5\wd\ptx\ptnext}
\def\ptnode#1{\ptnodeaux{\ptnodefn\ptnodestrut #1}}
\def\ptleaf#1{\ptnodeaux{\ptleaffn\ptleafstrut #1}}
\def\pthoradjust#1{\ifcase\ptn
  \or \pthadjbox{\ptm}{#1} \or \pthadjbox{\pta}{#1}
  \or \pthadjbox{\ptb}{#1} \or \pthadjbox{\ptc}{#1}
  \else \pterror{More than 3 daughters in (sub)tree}\fi}
\def\pthadjbox#1#2{\setbox#1=\hbox{\box#1\kern#2}}
\def\ptcons
 {\ifnum\ptn=0 \ptconsz 
  \else
    \ifpttri\ptconstri\else
      \ifcase\ptn \or\ptconsm\or\ptconsma\or\ptconsmab\or\ptconsmabc \fi \fi
    \ptax=\ptxx \advance\ptax-\ptmx \ptbx=0pt
    \ifdim\ptax<0pt \ptbx=-\ptax\ptax=0pt\ptxx=\ptmx \fi
    \setbox\ptx=\vtop{\hbox{\kern\ptax\box\ptm}\nointerlineskip
                      \hbox{\kern\ptbx\box\ptx}}\fi
  \global\ptxx=\ptxx\global\setbox\ptx=\box\ptx}
\def\ptavg#1#2#3{#1=#2\advance#1#3#1=0.5#1}     
\def\ptadv#1#2{\advance#1#2\advance#1\pthgap}   
\def\ptconsz{\ptxx=0pt \setbox\ptx=\vtop{}}     
\def\ptconsm{\ptxx=0pt 
  \setbox\ptx=\hbox{\ptedge{1}{0}{}{}}}         
\def\ptconsma                                   
 {\ptxx=\ptax\setbox\ptx=\vtop{
    \hbox{\ptedge{1}\ptax{}{}}\nointerlineskip
    \hbox{\box\pta}}}
\def\ptconsmab                                  
 {\ptadv\ptbx{\wd\pta}\ptavg\ptxx\ptax\ptbx
  \setbox\ptx=\vtop{
    \hbox{\ptedge{2}\ptax\ptbx{}}\nointerlineskip
    \hbox{\box\pta\kern\pthgap\box\ptb}}}
\def\ptconsmabc                                 
 {\ptadv\ptbx{\wd\pta}\ptadv\ptcx{\wd\pta}%
  \ptadv\ptcx{\wd\ptb}\ptavg\ptxx\ptax\ptcx
  \setbox\ptx=\vtop{
    \hbox{\ptedge{3}\ptax\ptbx\ptcx}\nointerlineskip
    \hbox{\box\pta\kern\pthgap\box\ptb\kern\pthgap\box\ptc}}}
\def\ptconstri                                  
 {\ifcase\ptn\or\setbox\pta\hbox{\kern2\pthgap}\or
  \or\setbox\pta=\hbox{\box\pta\kern\pthgap\box\ptb}
  \or\setbox\pta=\hbox{\box\pta\kern\pthgap\box\ptb\kern\pthgap\box\ptc}\fi
  \ptxx=0.5\wd\pta
  \setbox\ptx=\vtop{
    \hbox{\ptedge{0}{0}{\wd\pta}{}}\nointerlineskip
    \box\pta}}
\newcount\pted          
\newcount\ptedm         
\newcount\pteda         
\newcount\ptedb         
\newcount\ptedc         
\newcount\ptedl         
\newcount\ptedh         
\newcount\ptedhs        
\newcount\ptedvs        
\newcount\ptedtemp      
\def\ptedge#1#2#3#4{\pted=#1%
  \pteda=#2\ifcase\pted\ptedb=#3\or\or\ptedb=#3\or\ptedb=#3\ptedc=#4\fi
  \ptedm=\pteda\advance\ptedm\ifcase\pted\ptedb\or\pteda\or\ptedb\or\ptedc\fi
  \divide\ptedm by 2
  \ptedh=\ptvgap\ptedtemp=\ptedm\advance\ptedtemp-\pteda\divide\ptedtemp by 6
  \ifnum\ptedh<\ptedtemp\ptedh=\ptedtemp\fi
  \unitlength=1sp%
  \begin{picture}(0,\ptedh)
    \ifnum\pted=3 \ptedput\ptedc\fi
    \ifnum\pted=1 \else\ptedput\ptedb\fi
    \ptedput\pteda
    \ifnum\pted=0 \ptedbot\fi 
  \end{picture}}
\def\ptedput#1{\ptedl=#1\advance\ptedl-\ptedm
  \ifnum\ptedl>0 \ptedslope\else
    \ptedl=-\ptedl\ptedslope\ptedhs=-\ptedhs\fi
  \ifnum\ptedhs=0 \ptedl=\ptedh\fi
  \put(\ptedm,\ptedh){\line(\ptedhs,-\ptedvs){\ptedl}}}
\def\ptedbot
 {\ptedtemp=\ptedl\multiply\ptedtemp by \ptedvs\divide\ptedtemp by \ptedhs
  \ifnum\ptedtemp>0\ptedtemp=-\ptedtemp\fi \advance\ptedtemp-1
  \advance\ptedtemp\ptedh\multiply\ptedl by 2
  \put(\pteda,\ptedtemp){\line(1,0){\ptedl}}}
\def\ptedslope
 {\ifnum\ptedl>\ptedh\ptedhs=\ptedl\ptedvs=\ptedh
  \else              \ptedvs=\ptedl\ptedhs=\ptedh \fi
  \divide\ptedhs by 60 \divide\ptedvs by \ptedhs
  \ifnum \ptedvs <  5 \ptedvs=0 \ptedhs=1 \else
  \ifnum \ptedvs < 11 \ptedvs=1 \ptedhs=6 \else
  \ifnum \ptedvs < 13 \ptedvs=1 \ptedhs=5 \else
  \ifnum \ptedvs < 17 \ptedvs=1 \ptedhs=4 \else
  \ifnum \ptedvs < 22 \ptedvs=1 \ptedhs=3 \else
  \ifnum \ptedvs < 27 \ptedvs=2 \ptedhs=5 \else
  \ifnum \ptedvs < 33 \ptedvs=1 \ptedhs=2 \else
  \ifnum \ptedvs < 38 \ptedvs=3 \ptedhs=5 \else
  \ifnum \ptedvs < 42 \ptedvs=2 \ptedhs=3 \else
  \ifnum \ptedvs < 46 \ptedvs=3 \ptedhs=4 \else
  \ifnum \ptedvs < 49 \ptedvs=4 \ptedhs=5 \else
  \ifnum \ptedvs < 55 \ptedvs=5 \ptedhs=6 \else
                      \ptedvs=1 \ptedhs=1 
    \fi \fi \fi \fi \fi \fi \fi \fi \fi \fi \fi \fi
  \ifnum \ptedl < \ptedh
    \ptedtemp=\ptedhs \ptedhs=\ptedvs \ptedvs=\ptedtemp \fi}

\def\ptcatcodes
 {\catcode`(=\active \catcode`)=\active
  \catcode`.=\active \catcode``=\active
  \catcode`~=\active}
{\ptcatcodes
\gdef\ptactivechardefs
 {\ptcatcodes
  \def({\ptbeg\ignorespaces}
  \def){\ptend\ignorespaces}
  \def.##1.{\ptnode{##1}\ignorespaces} 
  \def`##1'{\ptleaf{##1}\ignorespaces}
  \def~{\pttritrue\ignorespaces}}
}

\def\bidx{b}

\def\measureset{[0,1]}

\def\Set{\mathbb{S}}

\def\oacc{\{}
\def\facc{\}}

\def\exp{e^}

\def\newlog{\mathfrak{L}}
\def\integ{\mathcal{T}}
\def\macrol{\mathcal{M}}
\def\netcrol{\mathcal{N}}

\def\dense{\mathfrak{D}}
\def\inputX{\mathfrak{F}}

\def\conv{\mathbb{C}}
\def\dsc{\mathbb{D}}
\def\conva{\mathbb{C}_{\textrm{L}}}
\def\dsca{\mathbb{D}_{\textrm{L}}}

\def\poolavg{\mathbb{F}}
\def\poolmax{\mathbb{G}}

  \def\minim{\textrm{Min}}
  \def\meanim{\textrm{Mean}}
  \def\medium{\textrm{Median}}
  \def\maxim{\textrm{Max}}

\def\perfo{\mathfrak{P}}

\def\ability{\mathcal{A}}

\def\measureset{[0,1]}

\def\bidx{b}

\def\prob{\mathbb{p}}
\def\qrob{\mathbb{q}}

\def\numparams{{\bold N}}
\def\euler{\mathfrak{e}}

\newtheorem{remark}{Remark}

{\hspace{\stretch{1}}%
\rule{1ex}{1ex}}

\tikzset{
  pics/.cd,
  disc/.style = {
    code = {
      \fill [white] ellipse [x radius = 1, y radius = 1/3];
      \path [left color = black!50, right color = black!50,
        middle color = black!25] 
        (-1+.05,-1.1) arc (180:360:1-.05 and 1/3-.05*1/3) -- cycle;
      \path [top color = black!25, bottom color = white] 
        (0,.05*1/3) ellipse [x radius = 1-.05, y radius = 1/3-.05*1/3];
      \path [left color = black!25, right color = black!25,
        middle color = white] (-1,0) -- (-1,-1) arc (180:360:1 and 1/3)
        -- (1,0) arc (360:180:1 and 1/3);
      \foreach \r in {225,315}
        \foreach \i [evaluate = {\s=30;}] in {0,2,...,30}
          \fill [black, fill opacity = 1/50] 
            (0,0) -- (\r+\s-\i:1 and 1/3) -- ++(0,-1) 
            arc (\r+\s-\i:\r-\s+\i:1 and 1/3) -- ++(0,1) -- cycle;
      \foreach \r in {45,135}
        \foreach \i [evaluate = {\s=30;}] in {0,2,...,30}
          \fill [black, fill opacity = 1/50] 
            (0,0) -- (\r+\s-\i:1 and 1/3) 
            arc (\r+\s-\i:\r-\s+\i:1 and 1/3)  -- cycle;
    }
  },
discgreen/.style = {
    code = {
      \fill [gray] ellipse [x radius = 1-0.05, y radius = 1/3];
      \path [left color = black!50, right color = black!50,
        middle color = black!25] 
        (-1+.05,-1.055) arc (180:360:1-.05 and 1/3-.05*1/3) -- cycle;
      \path [top color = black!25, bottom color = white] 
        (0,.05*1/3) ellipse [x radius = 1-.05, y radius = 1/3-.05*1/3];
      \path [left color = mygreen!25, right color = mygreen!25,
        middle color = mygreen] (-1,0) -- (-1,-1) arc (180:360:1 and 1/3)
        -- (1,0) arc (360:180:1 and 1/3);
      \foreach \r in {225,315}
        \foreach \i [evaluate = {\s=30;}] in {0,2,...,30}
          \fill [mygreen, fill opacity = 1/50] 
            (0,0) -- (\r+\s-\i:1 and 1/3) -- ++(0,-1) 
            arc (\r+\s-\i:\r-\s+\i:1 and 1/3) -- ++(0,1) -- cycle;
      \foreach \r in {45,135}
        \foreach \i [evaluate = {\s=30;}] in {0,2,...,30}
          \fill [mygreen, fill opacity = 1/50] 
            (0,0) -- (\r+\s-\i:1 and 1/3) 
            arc (\r+\s-\i:\r-\s+\i:1 and 1/3)  -- cycle;
    }
  },
discredish/.style = {
    code = {
      \fill [gray] ellipse [x radius = 1-0.05, y radius = 1/3];
      \path [left color = black!50, right color = black!50,
        middle color = black!25] 
        (-1+.05,-1.055) arc (180:360:1-.05 and 1/3-.05*1/3) -- cycle;
      \path [top color = black!25, bottom color = white] 
        (0,.05*1/3) ellipse [x radius = 1-.05, y radius = 1/3-.05*1/3];
      \path [left color = redish!25, right color = redish!25,
        middle color = redish] (-1,0) -- (-1,-1) arc (180:360:1 and 1/3)
        -- (1,0) arc (360:180:1 and 1/3);
      \foreach \r in {225,315}
        \foreach \i [evaluate = {\s=30;}] in {0,2,...,30}
          \fill [redish, fill opacity = 1/50] 
            (0,0) -- (\r+\s-\i:1 and 1/3) -- ++(0,-1) 
            arc (\r+\s-\i:\r-\s+\i:1 and 1/3) -- ++(0,1) -- cycle;
      \foreach \r in {45,135}
        \foreach \i [evaluate = {\s=30;}] in {0,2,...,30}
          \fill [redish, fill opacity = 1/50] 
            (0,0) -- (\r+\s-\i:1 and 1/3) 
            arc (\r+\s-\i:\r-\s+\i:1 and 1/3)  -- cycle;
    }
  },
discblue/.style = {
    code = {
      \fill [gray] ellipse [x radius = 1-0.05, y radius = 1/3];
      \path [left color = black!50, right color = black!50,
        middle color = black!25] 
        (-1+.05,-1.055) arc (180:360:1-.05 and 1/3-.05*1/3) -- cycle;
      \path [top color = black!25, bottom color = white] 
        (0,.05*1/3) ellipse [x radius = 1-.05, y radius = 1/3-.05*1/3];
      \path [left color = myblue!25, right color = myblue!25,
        middle color = myblue] (-1,0) -- (-1,-1) arc (180:360:1 and 1/3)
        -- (1,0) arc (360:180:1 and 1/3);
      \foreach \r in {225,315}
        \foreach \i [evaluate = {\s=30;}] in {0,2,...,30}
          \fill [myblue, fill opacity = 1/50] 
            (0,0) -- (\r+\s-\i:1 and 1/3) -- ++(0,-1) 
            arc (\r+\s-\i:\r-\s+\i:1 and 1/3) -- ++(0,1) -- cycle;
      \foreach \r in {45,135}
        \foreach \i [evaluate = {\s=30;}] in {0,2,...,30}
          \fill [myblue, fill opacity = 1/50] 
            (0,0) -- (\r+\s-\i:1 and 1/3) 
            arc (\r+\s-\i:\r-\s+\i:1 and 1/3)  -- cycle;
    }
  },
discmagenta/.style = {
    code = {
      \fill [gray] ellipse [x radius = 1-0.05, y radius = 1/3];
      \path [left color = black!50, right color = black!50,
        middle color = black!25] 
        (-1+.05,-1.055) arc (180:360:1-.05 and 1/3-.05*1/3) -- cycle;
      \path [top color = black!25, bottom color = white] 
        (0,.05*1/3) ellipse [x radius = 1-.05, y radius = 1/3-.05*1/3];
      \path [left color = magenta!25, right color = magenta!25,
        middle color = magenta] (-1,0) -- (-1,-1) arc (180:360:1 and 1/3)
        -- (1,0) arc (360:180:1 and 1/3);
      \foreach \r in {225,315}
        \foreach \i [evaluate = {\s=30;}] in {0,2,...,30}
          \fill [magenta, fill opacity = 1/50] 
            (0,0) -- (\r+\s-\i:1 and 1/3) -- ++(0,-1) 
            arc (\r+\s-\i:\r-\s+\i:1 and 1/3) -- ++(0,1) -- cycle;
      \foreach \r in {45,135}
        \foreach \i [evaluate = {\s=30;}] in {0,2,...,30}
          \fill [magenta, fill opacity = 1/50] 
            (0,0) -- (\r+\s-\i:1 and 1/3) 
            arc (\r+\s-\i:\r-\s+\i:1 and 1/3)  -- cycle;
    }
  },
discmycyan/.style = {
    code = {
      \fill [gray] ellipse [x radius = 1-0.05, y radius = 1/3];
      \path [left color = black!50, right color = black!50,
        middle color = black!25] 
        (-1+.05,-1.055) arc (180:360:1-.05 and 1/3-.05*1/3) -- cycle;
      \path [top color = black!25, bottom color = white] 
        (0,.05*1/3) ellipse [x radius = 1-.05, y radius = 1/3-.05*1/3];
      \path [left color = mycyan!25, right color = mycyan!25,
        middle color = mycyan] (-1,0) -- (-1,-1) arc (180:360:1 and 1/3)
        -- (1,0) arc (360:180:1 and 1/3);
      \foreach \r in {225,315}
        \foreach \i [evaluate = {\s=30;}] in {0,2,...,30}
          \fill [mycyan, fill opacity = 1/50] 
            (0,0) -- (\r+\s-\i:1 and 1/3) -- ++(0,-1) 
            arc (\r+\s-\i:\r-\s+\i:1 and 1/3) -- ++(0,1) -- cycle;
      \foreach \r in {45,135}
        \foreach \i [evaluate = {\s=30;}] in {0,2,...,30}
          \fill [mycyan, fill opacity = 1/50] 
            (0,0) -- (\r+\s-\i:1 and 1/3) 
            arc (\r+\s-\i:\r-\s+\i:1 and 1/3)  -- cycle;
    }
  },
discbrown/.style = {
    code = {
      \fill [gray] ellipse [x radius = 1-0.05, y radius = 1/3];
      \path [left color = black!50, right color = black!50,
        middle color = black!25] 
        (-1+.05,-1.055) arc (180:360:1-.05 and 1/3-.05*1/3) -- cycle;
      \path [top color = black!25, bottom color = white] 
        (0,.05*1/3) ellipse [x radius = 1-.05, y radius = 1/3-.05*1/3];
      \path [left color = brown!25, right color = brown!25,
        middle color = brown] (-1,0) -- (-1,-1) arc (180:360:1 and 1/3)
        -- (1,0) arc (360:180:1 and 1/3);
      \foreach \r in {225,315}
        \foreach \i [evaluate = {\s=30;}] in {0,2,...,30}
          \fill [brown, fill opacity = 1/50] 
            (0,0) -- (\r+\s-\i:1 and 1/3) -- ++(0,-1) 
            arc (\r+\s-\i:\r-\s+\i:1 and 1/3) -- ++(0,1) -- cycle;
      \foreach \r in {45,135}
        \foreach \i [evaluate = {\s=30;}] in {0,2,...,30}
          \fill [brown, fill opacity = 1/50] 
            (0,0) -- (\r+\s-\i:1 and 1/3) 
            arc (\r+\s-\i:\r-\s+\i:1 and 1/3)  -- cycle;
    }
  },
discgreenwide/.style = {
    code = {
      \fill [gray] ellipse [x radius = 1.5-0.05, y radius = 1/3];
      \path [left color = black!50, right color = black!50,
        middle color = black!25] 
        (-1.5+.05,-1.055) arc (180:360:1.5-.05 and 1/3-.05*1/3) -- cycle;
      \path [top color = black!25, bottom color = white] 
        (0,.05*1/3) ellipse [x radius = 1.5-.05, y radius = 1/3-.05*1/3];
      \path [left color = mygreen!25, right color = mygreen!25,
        middle color = mygreen] (-1.5,0) -- (-1.5,-1) arc (180:360:1.5 and 1/3)
        -- (1.5,0) arc (360:180:1.5 and 1/3);
      \foreach \r in {225,315}
        \foreach \i [evaluate = {\s=30;}] in {0,2,...,30}
          \fill [mygreen, fill opacity = 1/50] 
            (0,0) -- (\r+\s-\i:1.5 and 1/3) -- ++(0,-1) 
            arc (\r+\s-\i:\r-\s+\i:1.5 and 1/3) -- ++(0,1) -- cycle;
      \foreach \r in {45,135}
        \foreach \i [evaluate = {\s=30;}] in {0,2,...,30}
          \fill [mygreen, fill opacity = 1/50] 
            (0,0) -- (\r+\s-\i:1.5 and 1/3) 
            arc (\r+\s-\i:\r-\s+\i:1.5 and 1/3)  -- cycle;
    }
  },
discredishwide/.style = {
    code = {
      \fill [gray] ellipse [x radius = 1.5-0.05, y radius = 1/3];
      \path [left color = black!50, right color = black!50,
        middle color = black!25] 
        (-1.5+.05,-1.055) arc (180:360:1.5-.05 and 1/3-.05*1/3) -- cycle;
      \path [top color = black!25, bottom color = white] 
        (0,.05*1/3) ellipse [x radius = 1.5-.05, y radius = 1/3-.05*1/3];
      \path [left color = redish!25, right color = redish!25,
        middle color = redish] (-1.5,0) -- (-1.5,-1) arc (180:360:1.5 and 1/3)
        -- (1.5,0) arc (360:180:1.5 and 1/3);
      \foreach \r in {225,315}
        \foreach \i [evaluate = {\s=30;}] in {0,2,...,30}
          \fill [redish, fill opacity = 1/50] 
            (0,0) -- (\r+\s-\i:1.5 and 1/3) -- ++(0,-1) 
            arc (\r+\s-\i:\r-\s+\i:1.5 and 1/3) -- ++(0,1) -- cycle;
      \foreach \r in {45,135}
        \foreach \i [evaluate = {\s=30;}] in {0,2,...,30}
          \fill [redish, fill opacity = 1/50] 
            (0,0) -- (\r+\s-\i:1.5 and 1/3) 
            arc (\r+\s-\i:\r-\s+\i:1.5 and 1/3)  -- cycle;
    }
  },
discbluewide/.style = {
    code = {
      \fill [gray] ellipse [x radius = 1.5-0.05, y radius = 1/3];
      \path [left color = black!50, right color = black!50,
        middle color = black!25] 
        (-1.5+.05,-1.055) arc (180:360:1.5-.05 and 1/3-.05*1/3) -- cycle;
      \path [top color = black!25, bottom color = white] 
        (0,.05*1/3) ellipse [x radius = 1.5-.05, y radius = 1/3-.05*1/3];
      \path [left color = myblue!25, right color = myblue!25,
        middle color = myblue] (-1.5,0) -- (-1.5,-1) arc (180:360:1.5 and 1/3)
        -- (1.5,0) arc (360:180:1.5 and 1/3);
      \foreach \r in {225,315}
        \foreach \i [evaluate = {\s=30;}] in {0,2,...,30}
          \fill [myblue, fill opacity = 1/50] 
            (0,0) -- (\r+\s-\i:1.5 and 1/3) -- ++(0,-1) 
            arc (\r+\s-\i:\r-\s+\i:1.5 and 1/3) -- ++(0,1) -- cycle;
      \foreach \r in {45,135}
        \foreach \i [evaluate = {\s=30;}] in {0,2,...,30}
          \fill [myblue, fill opacity = 1/50] 
            (0,0) -- (\r+\s-\i:1.5 and 1/3) 
            arc (\r+\s-\i:\r-\s+\i:1.5 and 1/3)  -- cycle;
    }
  },
discmagentawide/.style = {
    code = {
      \fill [gray] ellipse [x radius = 1.5-0.05, y radius = 1/3];
      \path [left color = black!50, right color = black!50,
        middle color = black!25] 
        (-1.5+.05,-1.055) arc (180:360:1.5-.05 and 1/3-.05*1/3) -- cycle;
      \path [top color = black!25, bottom color = white] 
        (0,.05*1/3) ellipse [x radius = 1.5-.05, y radius = 1/3-.05*1/3];
      \path [left color = magenta!25, right color = magenta!25,
        middle color = magenta] (-1.5,0) -- (-1.5,-1) arc (180:360:1.5 and 1/3)
        -- (1.5,0) arc (360:180:1.5 and 1/3);
      \foreach \r in {225,315}
        \foreach \i [evaluate = {\s=30;}] in {0,2,...,30}
          \fill [magenta, fill opacity = 1/50] 
            (0,0) -- (\r+\s-\i:1.5 and 1/3) -- ++(0,-1) 
            arc (\r+\s-\i:\r-\s+\i:1.5 and 1/3) -- ++(0,1) -- cycle;
      \foreach \r in {45,135}
        \foreach \i [evaluate = {\s=30;}] in {0,2,...,30}
          \fill [magenta, fill opacity = 1/50] 
            (0,0) -- (\r+\s-\i:1.5 and 1/3) 
            arc (\r+\s-\i:\r-\s+\i:1.5 and 1/3)  -- cycle;
    }
  },
discmycyanwide/.style = {
    code = {
      \fill [gray] ellipse [x radius = 1.5-0.05, y radius = 1/3];
      \path [left color = black!50, right color = black!50,
        middle color = black!25] 
        (-1.5+.05,-1.055) arc (180:360:1.5-.05 and 1/3-.05*1/3) -- cycle;
      \path [top color = black!25, bottom color = white] 
        (0,.05*1/3) ellipse [x radius = 1.5-.05, y radius = 1/3-.05*1/3];
      \path [left color = mycyan!25, right color = mycyan!25,
        middle color = mycyan] (-1.5,0) -- (-1.5,-1) arc (180:360:1.5 and 1/3)
        -- (1.5,0) arc (360:180:1.5 and 1/3);
      \foreach \r in {225,315}
        \foreach \i [evaluate = {\s=30;}] in {0,2,...,30}
          \fill [mycyan, fill opacity = 1/50] 
            (0,0) -- (\r+\s-\i:1.5 and 1/3) -- ++(0,-1) 
            arc (\r+\s-\i:\r-\s+\i:1.5 and 1/3) -- ++(0,1) -- cycle;
      \foreach \r in {45,135}
        \foreach \i [evaluate = {\s=30;}] in {0,2,...,30}
          \fill [mycyan, fill opacity = 1/50] 
            (0,0) -- (\r+\s-\i:1.5 and 1/3) 
            arc (\r+\s-\i:\r-\s+\i:1.5 and 1/3)  -- cycle;
    }
  },
discbrownwide/.style = {
    code = {
      \fill [gray] ellipse [x radius = 1.5-0.05, y radius = 1/3];
      \path [left color = black!50, right color = black!50,
        middle color = black!25] 
        (-1.5+.05,-1.055) arc (180:360:1.5-.05 and 1/3-.05*1/3) -- cycle;
      \path [top color = black!25, bottom color = white] 
        (0,.05*1/3) ellipse [x radius = 1.5-.05, y radius = 1/3-.05*1/3];
      \path [left color = brown!25, right color = brown!25,
        middle color = brown] (-1.5,0) -- (-1.5,-1) arc (180:360:1.5 and 1/3)
        -- (1.5,0) arc (360:180:1.5 and 1/3);
      \foreach \r in {225,315}
        \foreach \i [evaluate = {\s=30;}] in {0,2,...,30}
          \fill [brown, fill opacity = 1/50] 
            (0,0) -- (\r+\s-\i:1.5 and 1/3) -- ++(0,-1) 
            arc (\r+\s-\i:\r-\s+\i:1.5 and 1/3) -- ++(0,1) -- cycle;
      \foreach \r in {45,135}
        \foreach \i [evaluate = {\s=30;}] in {0,2,...,30}
          \fill [brown, fill opacity = 1/50] 
            (0,0) -- (\r+\s-\i:1.5 and 1/3) 
            arc (\r+\s-\i:\r-\s+\i:1.5 and 1/3)  -- cycle;
    }
  },
discgreennormal/.style = {
    code = {
      \fill [gray] ellipse [x radius = 1.25-0.05, y radius = 1/3];
      \path [left color = black!50, right color = black!50,
        middle color = black!25] 
        (-1.25+.05,-1.055) arc (180:360:1.25-.05 and 1/3-.05*1/3) -- cycle;
      \path [top color = black!25, bottom color = white] 
        (0,.05*1/3) ellipse [x radius = 1.25-.05, y radius = 1/3-.05*1/3];
      \path [left color = mygreen!25, right color = mygreen!25,
        middle color = mygreen] (-1.25,0) -- (-1.25,-1) arc (180:360:1.25 and 1/3)
        -- (1.25,0) arc (360:180:1.25 and 1/3);
      \foreach \r in {225,315}
        \foreach \i [evaluate = {\s=30;}] in {0,2,...,30}
          \fill [mygreen, fill opacity = 1/50] 
            (0,0) -- (\r+\s-\i:1.25 and 1/3) -- ++(0,-1) 
            arc (\r+\s-\i:\r-\s+\i:1.25 and 1/3) -- ++(0,1) -- cycle;
      \foreach \r in {45,135}
        \foreach \i [evaluate = {\s=30;}] in {0,2,...,30}
          \fill [mygreen, fill opacity = 1/50] 
            (0,0) -- (\r+\s-\i:1.25 and 1/3) 
            arc (\r+\s-\i:\r-\s+\i:1.25 and 1/3)  -- cycle;
    }
  },
discredishnormal/.style = {
    code = {
      \fill [gray] ellipse [x radius = 1.25-0.05, y radius = 1/3];
      \path [left color = black!50, right color = black!50,
        middle color = black!25] 
        (-1.25+.05,-1.055) arc (180:360:1.25-.05 and 1/3-.05*1/3) -- cycle;
      \path [top color = black!25, bottom color = white] 
        (0,.05*1/3) ellipse [x radius = 1.25-.05, y radius = 1/3-.05*1/3];
      \path [left color = redish!25, right color = redish!25,
        middle color = redish] (-1.25,0) -- (-1.25,-1) arc (180:360:1.25 and 1/3)
        -- (1.25,0) arc (360:180:1.25 and 1/3);
      \foreach \r in {225,315}
        \foreach \i [evaluate = {\s=30;}] in {0,2,...,30}
          \fill [redish, fill opacity = 1/50] 
            (0,0) -- (\r+\s-\i:1.25 and 1/3) -- ++(0,-1) 
            arc (\r+\s-\i:\r-\s+\i:1.25 and 1/3) -- ++(0,1) -- cycle;
      \foreach \r in {45,135}
        \foreach \i [evaluate = {\s=30;}] in {0,2,...,30}
          \fill [redish, fill opacity = 1/50] 
            (0,0) -- (\r+\s-\i:1.25 and 1/3) 
            arc (\r+\s-\i:\r-\s+\i:1.25 and 1/3)  -- cycle;
    }
  },
discbluenormal/.style = {
    code = {
      \fill [gray] ellipse [x radius = 1.25-0.05, y radius = 1/3];
      \path [left color = black!50, right color = black!50,
        middle color = black!25] 
        (-1.25+.05,-1.055) arc (180:360:1.25-.05 and 1/3-.05*1/3) -- cycle;
      \path [top color = black!25, bottom color = white] 
        (0,.05*1/3) ellipse [x radius = 1.25-.05, y radius = 1/3-.05*1/3];
      \path [left color = myblue!25, right color = myblue!25,
        middle color = myblue] (-1.25,0) -- (-1.25,-1) arc (180:360:1.25 and 1/3)
        -- (1.25,0) arc (360:180:1.25 and 1/3);
      \foreach \r in {225,315}
        \foreach \i [evaluate = {\s=30;}] in {0,2,...,30}
          \fill [myblue, fill opacity = 1/50] 
            (0,0) -- (\r+\s-\i:1.25 and 1/3) -- ++(0,-1) 
            arc (\r+\s-\i:\r-\s+\i:1.25 and 1/3) -- ++(0,1) -- cycle;
      \foreach \r in {45,135}
        \foreach \i [evaluate = {\s=30;}] in {0,2,...,30}
          \fill [myblue, fill opacity = 1/50] 
            (0,0) -- (\r+\s-\i:1.25 and 1/3) 
            arc (\r+\s-\i:\r-\s+\i:1.25 and 1/3)  -- cycle;
    }
  },
discmagentanormal/.style = {
    code = {
      \fill [gray] ellipse [x radius = 1.25-0.05, y radius = 1/3];
      \path [left color = black!50, right color = black!50,
        middle color = black!25] 
        (-1.25+.05,-1.055) arc (180:360:1.25-.05 and 1/3-.05*1/3) -- cycle;
      \path [top color = black!25, bottom color = white] 
        (0,.05*1/3) ellipse [x radius = 1.25-.05, y radius = 1/3-.05*1/3];
      \path [left color = magenta!25, right color = magenta!25,
        middle color = magenta] (-1.25,0) -- (-1.25,-1) arc (180:360:1.25 and 1/3)
        -- (1.25,0) arc (360:180:1.25 and 1/3);
      \foreach \r in {225,315}
        \foreach \i [evaluate = {\s=30;}] in {0,2,...,30}
          \fill [magenta, fill opacity = 1/50] 
            (0,0) -- (\r+\s-\i:1.25 and 1/3) -- ++(0,-1) 
            arc (\r+\s-\i:\r-\s+\i:1.25 and 1/3) -- ++(0,1) -- cycle;
      \foreach \r in {45,135}
        \foreach \i [evaluate = {\s=30;}] in {0,2,...,30}
          \fill [magenta, fill opacity = 1/50] 
            (0,0) -- (\r+\s-\i:1.25 and 1/3) 
            arc (\r+\s-\i:\r-\s+\i:1.25 and 1/3)  -- cycle;
    }
  },
discmycyannormal/.style = {
    code = {
      \fill [gray] ellipse [x radius = 1.25-0.05, y radius = 1/3];
      \path [left color = black!50, right color = black!50,
        middle color = black!25] 
        (-1.25+.05,-1.055) arc (180:360:1.25-.05 and 1/3-.05*1/3) -- cycle;
      \path [top color = black!25, bottom color = white] 
        (0,.05*1/3) ellipse [x radius = 1.25-.05, y radius = 1/3-.05*1/3];
      \path [left color = mycyan!25, right color = mycyan!25,
        middle color = mycyan] (-1.25,0) -- (-1.25,-1) arc (180:360:1.25 and 1/3)
        -- (1.25,0) arc (360:180:1.25 and 1/3);
      \foreach \r in {225,315}
        \foreach \i [evaluate = {\s=30;}] in {0,2,...,30}
          \fill [mycyan, fill opacity = 1/50] 
            (0,0) -- (\r+\s-\i:1.25 and 1/3) -- ++(0,-1) 
            arc (\r+\s-\i:\r-\s+\i:1.25 and 1/3) -- ++(0,1) -- cycle;
      \foreach \r in {45,135}
        \foreach \i [evaluate = {\s=30;}] in {0,2,...,30}
          \fill [mycyan, fill opacity = 1/50] 
            (0,0) -- (\r+\s-\i:1.25 and 1/3) 
            arc (\r+\s-\i:\r-\s+\i:1.25 and 1/3)  -- cycle;
    }
  },
discbrownnormal/.style = {
    code = {
      \fill [gray] ellipse [x radius = 1.25-0.05, y radius = 1/3];
      \path [left color = black!50, right color = black!50,
        middle color = black!25] 
        (-1.25+.05,-1.055) arc (180:360:1.25-.05 and 1/3-.05*1/3) -- cycle;
      \path [top color = black!25, bottom color = white] 
        (0,.05*1/3) ellipse [x radius = 1.25-.05, y radius = 1/3-.05*1/3];
      \path [left color = brown!25, right color = brown!25,
        middle color = brown] (-1.25,0) -- (-1.25,-1) arc (180:360:1.25 and 1/3)
        -- (1.25,0) arc (360:180:1.25 and 1/3);
      \foreach \r in {225,315}
        \foreach \i [evaluate = {\s=30;}] in {0,2,...,30}
          \fill [brown, fill opacity = 1/50] 
            (0,0) -- (\r+\s-\i:1.25 and 1/3) -- ++(0,-1) 
            arc (\r+\s-\i:\r-\s+\i:1.25 and 1/3) -- ++(0,1) -- cycle;
      \foreach \r in {45,135}
        \foreach \i [evaluate = {\s=30;}] in {0,2,...,30}
          \fill [brown, fill opacity = 1/50] 
            (0,0) -- (\r+\s-\i:1.25 and 1/3) 
            arc (\r+\s-\i:\r-\s+\i:1.25 and 1/3)  -- cycle;
    }
  },
  disc bottom/.style = {
    code = {
      \foreach \i in {0,2,...,30}
        \fill [black, fill opacity = 0/60] (0,-1.1) 
          ellipse [x radius = 1+\i/40, y radius = 1/3+\i/60];
      \path pic {disc};
    }
  }
}

\author{Abdourrahmane M. {\sc ATTO}\,\thanks{ Email: \texttt{Abdourrahmane.Atto@univ-smb.fr} - Phone: +334 50 09 65 27 - Fax: +334 50 09 65 59 }             \,  
  \thanks{Support (contextual) from: 
  Multidisciplinary Institute in Artificial Intelligence (MIAI Cluster, ANR).   }                   \\                      {  Universit\'e Savoie Mont Blanc}       \\        {LISTIC, Polytech Annecy-Chamb\'ery, France}                    
  }

\usepackage[nonatbib,final]{neuripstyle}

\usepackage[utf8]{inputenc} 
\usepackage[T1]{fontenc}    
\usepackage{hyperref}       
\usepackage{url}            
\usepackage{booktabs}       
\usepackage{amsfonts}       
\usepackage{nicefrac}       
\usepackage{microtype}      
\usepackage{xcolor}         

\title{
Softlog-Softmax Layers and Divergences Contribute to a Computationally Dependable Ensemble Learning 
}

\begin{document}

\maketitle

\begin{abstract}
The paper proposes a 4-step process for highlighting that softlog-softmax cascades can improve both consistency and dependability of the next generation ensemble learning systems.  The first process is anatomical in nature: the target ensemble model under consideration is composed by canonical elements relating to the definition of a convolutional frustum. No a priori is considered in the choice of canonical forms. Diversity is the main criterion for selecting these forms. It is shown that the more complex the problem, the more useful this ensemble diversity is. The second process is physiological and relates to neural engineering: a softlog is derived to both make weak logarithmic operations consistent and lead, through multiple softlog-softmax layers, to intermediate decisions in the sense of respecting the same class logic as that faced by the output layer. The third process concerns neural information theory: softlog-based entropy and divergence are proposed for the sake of constructing information measures yielding consistent values on closed intervals. These information measures are used to determine the relationships between individual and sub-community decisions in frustum diversity-based ensemble learning. The concluding process addresses the derivation of an informative performance tensor for the purpose of a reliable ensemble evaluation.
\end{abstract}

\section{Introduction}


Convolutional Neural Network (CNN) models are ever-growing in both deepness and ensembles of parallel branches, which is somewhat disconcerting.  
Indeed, an insect like the dragonfly has approximately 250 thousand neurons \cite{dragonfly21}, which are sufficient for analyzing a dense optical flow together with possessing a selective visual attention that allows him to achieve high performance in high speed and with low computing power.

Neural Architecture Search  (NAS, see \cite{10043644}, \cite{10516268} among other references) can, without too much doubt but with a great deal of computational effort, make finding a frugal model possible for almost every standard circumstance. However, changing the model for each circumstance is not a viable alternative for a wide variety of future unsolved problems. Thus, we think that the future research will rather prioritize predefined ensemble models, even if it means reconfiguring their neural connections when necessary, in the context of a new challenge. After all, the same major statistical/geometrical invariances prevail almost everywhere, and it is therefore legitimate that very little effort will be made on reconfiguration to make a model performant (frugality deployed on transfer and adaptation rather than on the plurality of architectures per problem).
This is why the case study proposed in Section \ref{sec frustums} is a wide width ensemble model constructed by multiple integrations on diverse canonical 3-depth convolutive frustum shapes.

Cross-entropy,  a beloved training objective, raises another challenge. Its best conditioner on discrete and finite feature sets is the softmax operator. But on the one hand, softmax output, consisting in values derived from an exponential over a sum of exponential values, appears very sparse in general (far from a smooth probability distribution), meaning that cross-entropy-based training can fail when computing the logarithm of a substantially numerical `0 probability' for large-scale input features or large numbers of output categories.
On the other hand, related to ensemble learning, seeking an optimal softmax integrator operating on ensemble softmax decisions can deviate significantly from the input algebra domain (exponentiation on exponentially distributed values), causing a failure during training.
For decision state aggregation, applying a logarithm on the softmax probabilities to be integrated yields a feature set that is more regular and favorable to weighted combinations. However, the issue is then numerical approximation as highlighted just above for the usage of the logarithm for instance in cross-entropy. This is, incidentally, why the ensemble learning literature relocates the aggregation stage to some hidden features preceding the softmax outputs. 
To address these last problems associated with the logarithm issue, we propose in Section \ref{sec softlog}, the design of a safe log-layer associated with a softlog operator and address the use of this layer in multiple softlog-softmax integration schemes.

A Class Activation Map (CAM) is a helpful tool for tentative explanations of the consistency of a hidden series of features/layers to a decision, see the abundant literature on the domain:
\cite{8269992}, 
\cite{8004993}, 
\cite{9174973}, 
\cite{9996553}, 
\cite{10120949}, 
\cite{9873854}, 
\cite{10234096}, 
\cite{9353268}, 
\cite{9961149}, 
\cite{10381766}.
The problem at hand with CAMs is to systematically investigate and analyze the behavior of convolution outputs with respect to various inputs and the corresponding categories, including but not limited to the initial training data, see references given above, among others.
CAMs are computed  \emph{a posteriori} of the training in this context, but they can also be used at the training step to condition the learning \cite{bralet2025cviu}, \cite{camTrain} or to augment data \cite{camDataAugment} in some applications where labeled data are difficult to collect.
But unlike a standard monolithic CNN, an ensemble model is usually highly parallel. As a result, the use of CAMs is likely to return too many parallel and series of attributes whose impact on the final decision is not necessarily guaranteed. To overcome this difficulty, and given that the softlog opens up new divergence perspectives by virtue of its safe nature when composed with a softmax, we propose in Section \ref{sec sld}, an alternative way of evaluating controversy in a decision. This evaluation will be based on softlog divergences  that are distributed on closed intervals. These measures will allow determining the roles played by individuals and sub-communities in the ensemble learning framework considered.

How much time do we spend retraining a model ? In the end, we report only the maximum performance! We can't subsequently ensure recapturing this performance if the programming environment upgrades its random number generator or its optimization solver \cite{sgdm2007}, \cite{sgdm2012}, \cite{sgdm2013}, especially if the objective - usually a function of several million variables - has a non-trivial shape. 
Section \ref{sec perfotensor} develops an approach which consists in first characterizing the model certification by providing its performance tensor from case studies which are simple in terms of computational complexity, then eventually report a single/balanced performance value deduced from the more complex problem (time is often lacking at this stage to perform numerous trials).

Section \ref{sec expe} is an illustrative relevance of a 3-depth ensemble model, provided with softlog-softmax layers, to learn  CAPTCHA\footnote{Completely Automated Public Turing test to tell Computers and Humans Apart \cite{captcha1}.} processed alphabet letters, where the softlog-softmax layering approach allows identifying the most significant contributory elements with remarkably reduced effort.

\begin{figure*}[!htp]
\centering
\begin{tabular}{@{}c@{~}||@{~}c@{}}\hline\hline
\includegraphics[width=0.41\textwidth]{MIAI_USMB_encoded_decoded_Page_2.pdf}
&
\includegraphics[width=0.41\textwidth]{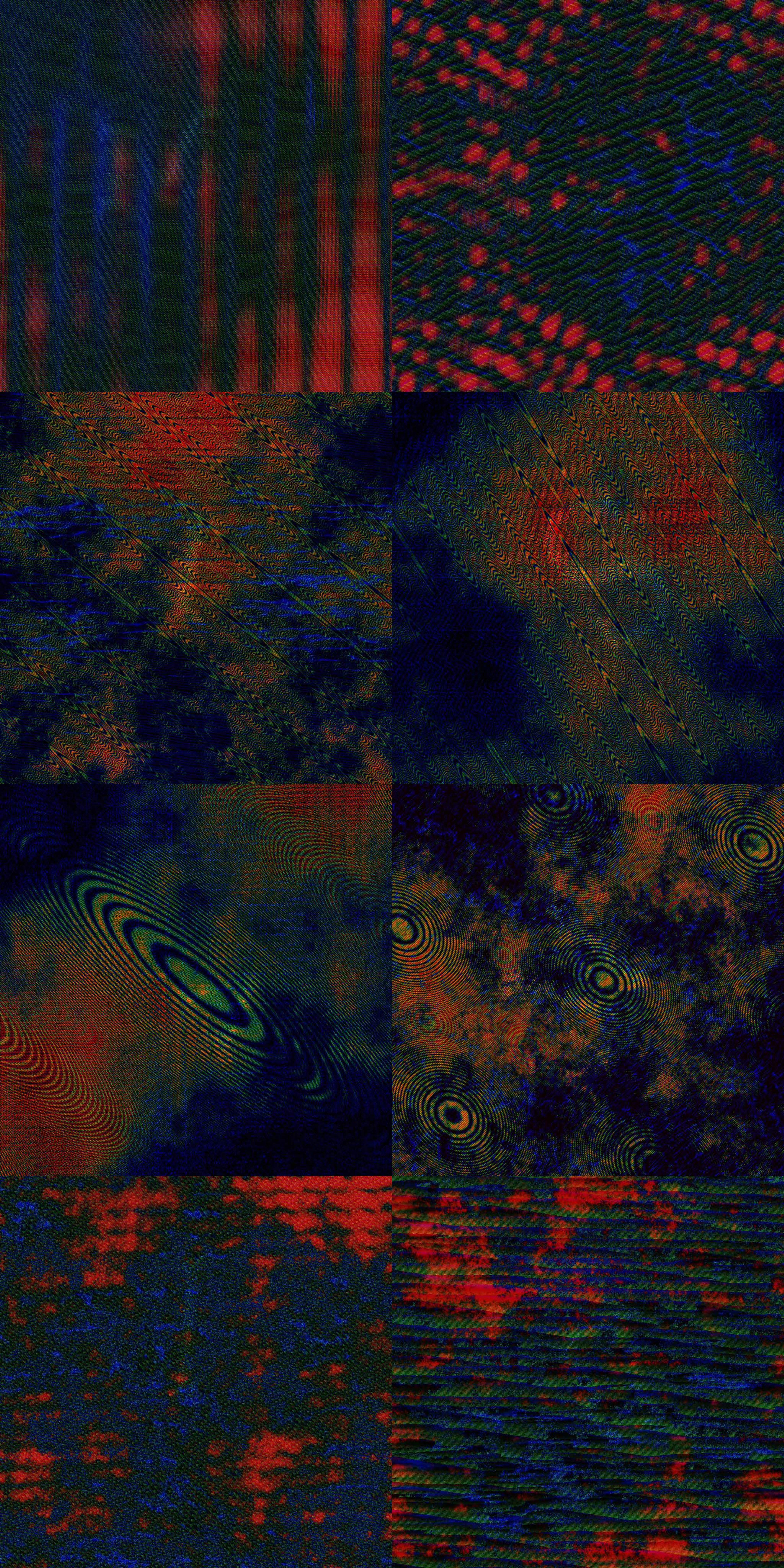}
\end{tabular}
\caption{ 
 Binary (left) and Encrypted SVA (right) geometrons ``M-U-I-S-A-M-I-B'' from \cite{atto24geometrons}.
} 
\label{fig state1images}
\end{figure*}

 \section{
 Operating ensemble learning system
 }\label{sec frustums}
 
\subsection{Visual geometron recognition problem}\label{ssec geometrons}

For all the experiments, we will use recognition scenarios associated with the ``geometrons'' \cite{atto24geometrons} dataset comprising both simple and very complex classification issues. 
In this dataset composed by 143 shape classes, geometrons are given either as binary or encrypted symbols. Encrypted symbols implied embedding a binary symbol in an intense visual texture stream, with the aim of investigating the limits of artificial visual attention to capture known or unknown, but ghost or buried shapes. Two levels of encryption have been used: the first level is called of \emph{Super-Visual Attention} (SVA) whereas the second level is associated with \emph{Ultra-Visual Attention} (UVA).
Binary and  encrypted geometron examples are given in Figure \ref{fig state1images}.
Specifically, we will focus on alphabet elements (26 classes; 119600 examples; any example is a letter given as binary or encrypted) from which several experiments will be set, from simple to more complex geometron recognition issues. We adopt the following distribution of geometron examples per category:
{Training samples}: 3519;  
{Validation samples}: 621;     
{Test samples}: 460.
The experimental setups used are the following. 
 \red{Experiment \#1: IO binary, classes are \{I, O\}};
 \myblue{Experiment \#2: BDOQ binary, classes are \{B, D, O, Q\}};
  \magenta{Experiment \#3: IO-SVA encrypted, classes are \{Encrypted[I], Encrypted[O]\}}.

\subsection{
DRAGONFLY: Determinants of Representational Architectures and Generational Objectives from Neural Frustum Learning Yields
}\label{sseceqprobs1to12}

A sequence of convolution layers is called a convolution frustum if its sizing specification, determined by the number of convolution filters per layer, is representative of a collection of frustum section sizes.
In this respect, the geometry of the layers (convolution branch) is a structure resembling a frustum.
We will consider both spatial 2D Convolution (2D-C) and Depthwise Separable Convolution (DSC) frustums. 
It is worth highlighting that the frustum definition adopted is independent of the spatial and channel sizes of the convolution filters. These spatial/channel informations refer to a convolution density criteria inside a given frustum shape. 
As a consequence, the cylindrical frustum terminology for instance will remain valid when the number of convolution filters remains constant whereas the frustum density increases from layer to layer in the given frustum. 

The DRAGONFLY model $\netcrol_{17}$  described by Figure \ref{gen-model} and Table \ref{tablesmodelsdragon0} integrates 12 networks associated with 6 elementary convolution frustums, respectively in 2D-C and DSC based frameworks.
The 6 basic frustums used are associated with cylindrical ($\netcrol_{1}, \netcrol_{2}$, $\netcrol_{7}, \netcrol_{8}$), contractive ($\netcrol_{3}, \netcrol_{9}$), hyperbolic ($\netcrol_{4}, \netcrol_{10}$), ovoid ($\netcrol_{5}, \netcrol_{11}$) and expansive ($\netcrol_{6}, \netcrol_{12}$) shapes.
 Note that any classifier $ \netcrol_{\bidx} $ given in Table \ref{tablesmodelsdragon0} is constructed from an independent DCN matrix $\dense^{[\bidx]} $ 
which is processed by a softmax operator to yield the output $\prob^{[\bidx]}$ of the elementary frustum-based network.

\begin{landscape}%
\begin{figure*}[!htp]
\vspace*{-5cm}
\hspace*{-1.5cm}
\centering
{\scriptsize
 \begin{tikzpicture}[scale=.3]
  \path 
  (-10,0) pic {discmagentawide} 
  (-15,-14-6+4) pic {discblue} %
  (-15,-14-3+4) pic {discredishnormal} %
  (-15,-14+4) pic {discmagentawide} %
  (-5,-14-6) pic {discmagentawide} 
  (-5,-14-3) pic {discblue} %
  (-5,-14) pic {discmagentawide} 
  (5,-14-6) pic {discblue} %
  (5,-14-3) pic {discmagentawide} 
  (5,-14) pic {discblue} %
  (15,-14-6+4) pic {discmagentawide} %
  (15,-14-3+4) pic {discredishnormal} %
  (15,-14+4) pic {discblue} %
  (10,-3) pic {discblue}   %
  (10,-0) pic {discblue}   %
  (10,3) pic {discblue}  %
  (-10+44,0) pic {discgreenwide} 
  (-15+44,-14-6+4) pic {discmycyan} %
  (-15+44,-14-3+4) pic {discbrownnormal} %
  (-15+44,-14+4) pic {discgreenwide} %
  (-5+44,-14-6) pic {discgreenwide} 
  (-5+44,-14-3) pic {discmycyan} %
  (-5+44,-14) pic {discgreenwide} 
  (5+44,-14-6) pic {discmycyan} %
  (5+44,-14-3) pic {discgreenwide} 
  (5+44,-14) pic {discmycyan} %
  (15+44,-14-6+4) pic {discgreenwide} %
  (15+44,-14-3+4) pic {discbrownnormal} %
  (15+44,-14+4) pic {discmycyan} %
  (10+44,-3) pic {discmycyan}   %
  (10+44,-0) pic {discmycyan}   %
  (10+44,3) pic {discmycyan};   %
\draw (-10,-3)  node (aa) {$\macrol_{1}$}; 
\draw (10,-0) node (aa) {$\macrol_{2,1}$}; 
\draw (10,-3) node (aa) {$\macrol_{2,2}$}; 
\draw (10,-3-3) node (aa) {$\macrol_{2,3}$}; 
\draw (-15,-14-3+4)  node (aa) {$\macrol_{3,1}$}; 
\draw (-15,-14-3+4-3)  node (aa) {$\macrol_{3,2}$}; 
\draw (-15,-14-3+4-6)  node (aa) {$\macrol_{3,3}$}; 
\draw (-5,-14-3)  node (aa) {$\macrol_{4,1}$}; 
\draw (-5,-14-3-3)  node (aa) {$\macrol_{4,2}$}; 
\draw (-5,-14-3-6)  node (aa) {$\macrol_{4,3}$}; 
\draw (5,-14-3)  node (aa) {$\macrol_{5,1}$}; 
\draw (5,-14-3-3)  node (aa) {$\macrol_{5,2}$}; 
\draw (5,-14-3-6)  node (aa) {$\macrol_{5,3}$}; 
\draw (15,-14-3+4)   node (aa) {$\macrol_{6,1}$}; 
\draw (15,-14-3+4-3)   node (aa) {$\macrol_{6,2}$}; 
\draw (15,-14-3+4-6)   node (aa) {$\macrol_{6,3}$}; 
\draw (-10+44,-3)  node (aa) {$\macrol_{7}$}; 
\draw (10+44,-0)  node (aa) {$\macrol_{8,1}$}; 
\draw (10+44,-0-3)  node (aa) {$\macrol_{8,2}$}; 
\draw (10+44,-0-6)  node (aa) {$\macrol_{8,3}$}; 
\draw (-15+44,-14-3+4)  node (aa) {$\macrol_{9,1}$}; 
\draw (-15+44,-14-3+4-3)  node (aa) {$\macrol_{9,2}$}; 
\draw (-15+44,-14-3+4-6)  node (aa) {$\macrol_{9,3}$}; 
\draw (-5+44,-14-3)  node (aa) {$\macrol_{10,1}$}; 
\draw (-5+44,-14-3-3)  node (aa) {$\macrol_{10,2}$}; 
\draw (-5+44,-14-3-6)  node (aa) {$\macrol_{10,3}$}; 
\draw (5+44,-14-3)  node (aa) {$\macrol_{11,1}$}; 
\draw (5+44,-14-3-3)  node (aa) {$\macrol_{11,2}$}; 
\draw (5+44,-14-3-6)  node (aa) {$\macrol_{11,3}$}; 
\draw (15+44,-14-3+4)  node (aa) {$\macrol_{12,1}$}; 
\draw (15+44,-14-3+4-3)  node (aa) {$\macrol_{12,2}$}; 
\draw (15+44,-14-3+4-6)  node (aa) {$\macrol_{12,3}$};

\draw (-10,-3+4)  node (aa) {$\!\!\!\!\!\inputX$ {\rotatebox{-90}{$\dashrightarrow$}}}; 
\draw (10,-0+4) node (aa) {$\!\!\!\!\!\inputX$ {\rotatebox{-90}{$\dashrightarrow$}}}; 
\draw (-15,-14-3+4+4)  node (aa) {$\!\!\!\!\!\inputX$ {\rotatebox{-90}{$\dashrightarrow$}}}; 
\draw (-5,-14-3+4)  node (aa) {$\!\!\!\!\!\inputX$ {\rotatebox{-90}{$\dashrightarrow$}}}; 
\draw (5,-14-3+4)  node (aa) {$\!\!\!\!\!\inputX$ {\rotatebox{-90}{$\dashrightarrow$}}}; 
\draw (15,-14-3+4+4)   node (aa) {$\!\!\!\!\!\inputX$ {\rotatebox{-90}{$\dashrightarrow$}}}; 
\draw (-10+44,-3+4)  node (aa) {$\!\!\!\!\!\inputX$ {\rotatebox{-90}{$\dashrightarrow$}}}; 
\draw (10+44,-0+4)  node (aa) {$\!\!\!\!\!\inputX$ {\rotatebox{-90}{$\dashrightarrow$}}}; 
\draw (-15+44,-14-3+4+4)  node (aa) {$\!\!\!\!\!\inputX$ {\rotatebox{-90}{$\dashrightarrow$}}}; 
\draw (-5+44,-14-3+4)  node (aa) {$\!\!\!\!\!\inputX$ {\rotatebox{-90}{$\dashrightarrow$}}}; 
\draw (5+44,-14-3+4)  node (aa) {$\!\!\!\!\!\inputX$ {\rotatebox{-90}{$\dashrightarrow$}}}; 
\draw (15+44,-14-3+4+4)  node (aa) {$\!\!\!\!\!\inputX$ {\rotatebox{-90}{$\dashrightarrow$}}};

 \blue{	
\draw (-10,0-4.5)  node (Mbottom) {$\rotatebox{-90}{$\dashrightarrow$}$}; 
\draw (-15,-14-6+4-4.5)  node (Cbottom) {$\rotatebox{-90}{$\dashrightarrow$}$}; 
\draw (-5,-14-6-4.5)  node (Kbottom) {$\rotatebox{-90}{$\dashrightarrow$}$}; 
\draw (5,-14-6-4.5)  node (Ebottom) {$\rotatebox{-90}{$\dashrightarrow$}$}; 
\draw (15,-14-6+4-4.5)  node (Obottom) {$\rotatebox{-90}{$\dashrightarrow$}$}; 
\draw (10,-3-4.5)  node (Fbottom) {$\rotatebox{-90}{$\dashrightarrow$}$}; 
\draw (0,3)  node (MFinteg) {\ovalbox{$\integ_{1,1}$}}; 
\draw (0,-9)  node (OCKEinteg) {\ovalbox{$\integ_{2,1}$}}; 
\draw[->> , line width=0.4mm] (Mbottom) .. controls (0,-8) and (0,1) .. node[below,pos=.57] {} (MFinteg);
\draw[->> , line width=0.4mm] (Fbottom) .. controls (0,-8) and (0,1) .. node[below,pos=.57] {} (MFinteg);
\draw[-> , line width=0.4mm] (Cbottom) .. controls (-4,-20) and (-18,-8) .. node[below,pos=.57] {} (OCKEinteg);
\draw[->> , line width=0.4mm] (Kbottom) .. controls (0,-20) and (0,-12) .. node[below,pos=.57] {} (OCKEinteg);
\draw[->> , line width=0.4mm] (Ebottom) .. controls (0,-20) and (0,-12) .. node[below,pos=.57] {} (OCKEinteg);
\draw[-> , line width=0.4mm] (Obottom) .. controls (2,-20) and (21,-8) .. node[below,pos=.57] {} (OCKEinteg);
}
 \redish{	
\draw (-10+44,0-4.5)  node (MbottomDS) {$\rotatebox{-90}{$\dashrightarrow$}$}; 
\draw (-15+44,-14-6+4-4.5)  node (CbottomDS) {$\rotatebox{-90}{$\dashrightarrow$}$}; 
\draw (-5+44,-14-6-4.5)  node (KbottomDS) {$\rotatebox{-90}{$\dashrightarrow$}$}; 
\draw (5+44,-14-6-4.5)  node (EbottomDS) {$\rotatebox{-90}{$\dashrightarrow$}$}; 
\draw (15+44,-14-6+4-4.5)  node (ObottomDS) {$\rotatebox{-90}{$\dashrightarrow$}$}; 
\draw (10+44,-3-4.5)  node (FbottomDS) {$\rotatebox{-90}{$\dashrightarrow$}$}; 
\draw (0+44,3)  node (MFintegDS) {\ovalbox{$\integ_{1,2}$}}; 
\draw (0+44,-9)  node (OCKEintegDS) {\ovalbox{$\integ_{2,2}$}}; 
\draw[->> , line width=0.4mm] (MbottomDS) .. controls (0+44,-8) and (0+44,1) .. node[below,pos=.57] {} (MFintegDS);
\draw[->> , line width=0.4mm] (FbottomDS) .. controls (0+44,-8) and (0+44,1) .. node[below,pos=.57] {} (MFintegDS);
\draw[-> , line width=0.4mm] (CbottomDS) .. controls (-4+44,-20) and (-18+44,-8) .. node[below,pos=.57] {} (OCKEintegDS);
\draw[->> , line width=0.4mm] (KbottomDS) .. controls (0+44,-20) and (0+44,-12) .. node[below,pos=.57] {} (OCKEintegDS);
\draw[->> , line width=0.4mm] (EbottomDS) .. controls (0+44,-20) and (0+44,-12) .. node[below,pos=.57] {} (OCKEintegDS);
\draw[-> , line width=0.4mm] (ObottomDS) .. controls (2+44,-20) and (21+44,-8) .. node[below,pos=.57] {} (OCKEintegDS);
}
\draw (22,-28.5)  node (Integrator) {\ovalbox{$\integ_{3}$}}; 
\draw[->> , line width=0.4mm] (MbottomDS) .. controls (20,-4) and (22,-23.5) .. node[below,pos=.57] {} (Integrator);
\draw[->> , line width=0.4mm] (MFintegDS) .. controls (20,8) and (22,-23.5) .. node[below,pos=.57] {} (Integrator);
\draw[->> , line width=0.4mm] (FbottomDS) .. controls (4,-2) and (28,-23.5) .. node[below,pos=.57] {} (Integrator);
\draw[->> , line width=0.4mm] (OCKEintegDS) .. controls (20,2) and (18,-23.5) .. node[below,pos=.57] {} (Integrator);
\draw[->> , line width=0.4mm] (Mbottom) .. controls (0,-15) and (32,4) .. node[below,pos=.57] {} (Integrator);
\draw[->> , line width=0.4mm] (MFinteg) .. controls (13,-24) and (22,10) .. node[below,pos=.57] {} (Integrator);
\draw[->> , line width=0.4mm] (Fbottom) .. controls (20,-7) and (22,10) .. node[below,pos=.57] {} (Integrator);
\draw[->> , line width=0.4mm] (OCKEinteg) .. controls (22,-18) and (22,25) .. node[below,pos=.57] {} (Integrator);
\draw[->> , line width=0.4mm] (CbottomDS) .. controls (30,-30.5) and (22,-32.5) .. node[below,pos=.57] {} (Integrator);
\draw[->> , line width=0.4mm] (KbottomDS) .. controls (40,-30.5) and (22,-32.5) .. node[below,pos=.57] {} (Integrator);
\draw[->> , line width=0.4mm] (EbottomDS) .. controls (50,-30.5) and (22,-32.5) .. node[below,pos=.57] {} (Integrator);
\draw[->> , line width=0.4mm] (ObottomDS) .. controls (60,-30.5) and (22,-32.5) .. node[below,pos=.57] {} (Integrator);
\draw[->> , line width=0.4mm] (Cbottom) .. controls (-15,-30.5) and (22,-32.5) .. node[below,pos=.57] {} (Integrator);
\draw[->> , line width=0.4mm] (Kbottom) .. controls (-5,-30.5) and (22,-32.5) .. node[below,pos=.57] {} (Integrator);
\draw[->> , line width=0.4mm] (Ebottom) .. controls (5,-30.5) and (22,-32.5) .. node[below,pos=.57] {} (Integrator);
\draw[->> , line width=0.4mm] (Obottom) .. controls (13,-30.5) and (22,-32.5) .. node[below,pos=.57] {} (Integrator);
\end{tikzpicture}
}
\caption{ 
{DRAGONFLY $\netcrol_{17}$ frustum conovolution structuring:
Elements $\integ_{1,1}$ and $\integ_{1,2}$ (parent's community);  $\integ_{2,1}$ and $\integ_{2,2}$(children community); $\integ_{3}$ (multi-generational).
Canonical network $\netcrol_{k}$ has convolution frustum size $\macrol_{k}$ with:
$\macrol_{1} = \macrol_{7} = \{128\}$: cylindric-wide frustum; 
$\macrol_{2} = \macrol_{8} = \{32, 32, 32\}$: cylindric-elongated frustum; 
$\macrol_{3} = \macrol_{9} = \{96,64,32\}$: contractive frustum; 
$\macrol_{4} = \macrol_{10} = \{80,48,80\}$: hyperbolic frustum; 
$\macrol_{5} = \macrol_{11} = \{48,80,48\}$: ovoid frustum; 
$\macrol_{6} = \macrol_{12} = \{32,64,96\}$: expansive frustum. 
} 
}
\label{gen-model}
\end{figure*}%
\end{landscape}



\begin{landscape}%
{
\newcolumntype{t}{>{\color{black}\columncolor{brown!12}[.5\tabcolsep]}c}  
\newcolumntype{s}{>{\color{black}\columncolor{red!12}[.5\tabcolsep]}c}  
\newcolumntype{o}{>{\color{black}\columncolor{magenta!12}[.5\tabcolsep]}c}  
\newcolumntype{m}{>{\color{black}\columncolor{magenta!24}[.5\tabcolsep]}c}  
\newcolumntype{n}{>{\color{black}\columncolor{red!24}[.5\tabcolsep]}c}  
\newcolumntype{y}{>{\color{black}\columncolor{orange!50}[.5\tabcolsep]}c}  
\newcolumntype{h}{>{\color{black}\columncolor{cyan!12}[.5\tabcolsep]}c}  
\newcolumntype{f}{>{\color{black}\columncolor{cyan!24}[.5\tabcolsep]}c}  
\newcolumntype{e}{>{\color{black}\columncolor{cyan!36}[.5\tabcolsep]}c}  
\newcolumntype{d}{>{\color{black}\columncolor{blue!12}[.5\tabcolsep]}c}  
\newcolumntype{b}{>{\color{black}\columncolor{blue!24}[.5\tabcolsep]}c}  
\newcolumntype{a}{>{\color{black}\columncolor{blue!36}[.5\tabcolsep]}c}  
\newcolumntype{g}{>{\columncolor[gray]{.95}[.5\tabcolsep]}c}  
\setlength{\arrayrulewidth}{0.35mm}
\begin{table*}[!htp]
\centering
\caption{
DRAGONFLY $\netcrol_{17}$ operating on canonical CNNs $\left\{ \netcrol_{k}, k=1, 2, \ldots, 12\right\} $ associated respectively with frustums of distribution shapes given as follows.
$\netcrol_{1}$ \& $\netcrol_{7}$: cylindric-wide; 
$\netcrol_{2}$ \& $\netcrol_{8}$: cylindric-elongated; 
$\netcrol_{3}$ \& $\netcrol_{9}$: contractive; 
$\netcrol_{4}$ \& $\netcrol_{10}$: hyperbolic; 
$\netcrol_{5}$ \& $\netcrol_{11}$: ovoid; 
$\netcrol_{6}$ \& $\netcrol_{12}$: expansive (reverse of contractive). 
Specifically in this table, we have used the following notations.
$\conva[N]$: $N$ spatial 2D convolution layers, each with filter size $(L, L)$.
$\dsca[N]$: Depthwise separable convolution layer composed by channelwise spatial convolutions with filter size $(L, L)$ and followed by $N$ pointwise convolutions.  The stride is fixed to 1 for all convolution operations.
$\poolavg_{M}[\downarrow\! N]$ indicates an average-pooling operation with respect to a spatial size $(M,M)$ and a stride $(N, N)$.
Similarly, $\poolmax_{M}[\downarrow\! N]$ indicates a max-pooling operation with respect to a spatial size $(M,M)$ and a stride $(N, N)$.
$K$ is the number of output categories for the classification process.
} 
\begin{scriptsize}
\begin{tabular}{|||@{~}t@{~} | @{~}s@{~} ||   @{~}o@{~} | @{~}m@{~} | @{~}n@{~} | @{~}y@{~}                      |||                  @{~}h@{~} | @{~}f@{~} ||    @{~}e@{~} | @{~}d@{~} | @{~}b@{~} | @{~}a@{~}                 ||}\hline\hline
  \multicolumn{12}{||c||}{
  Models 2D-C $\netcrol_{1}, \netcrol_{2}, \ldots, \netcrol_{6}$ and DSC $\netcrol_{7}, \netcrol_{8}, \ldots, \netcrol_{12}$ 
  }  \\\hline\hline
  \multicolumn{6}{||g||}{2D-C (standard 2D Convolutions)}  &  \multicolumn{6}{c||}{DSC (Depthwise Separable Convolutions)} \\\hline
 \multicolumn{2}{||g|}{Generation \#1}  &  \multicolumn{4}{c||}{Generation \#2}  &   \multicolumn{2}{g|}{Generation \#1}  &   \multicolumn{4}{c||}{Generation \#2}    \\\hline
$\netcrol_{1}$       &     $\netcrol_{2}$    &     $\netcrol_{3}$    &     $\netcrol_{4}$    &     $\netcrol_{5}$    &     $\netcrol_{6}$    &     $\netcrol_{7}$    &     $\netcrol_{8}$    &     $\netcrol_{9}$    &     $\netcrol_{10}$    &     $\netcrol_{11}$    &     $\netcrol_{12}$       \\   \hline\hline
                   $\conv_{3}[128]$       &    $\conv_{3}[32]$     &     $\conv_{3}[96]$     &     $\conv_{3}[80]$     &    $\conv_{3}[48]$     &     $\conv_{3}[32]$     &     $\dsc_{3}[128]$       &    $\dsc_{3}[32]$     &     $\dsc_{3}[96]$     &     $\dsc_{3}[80]$     &    $\dsc_{3}[48]$     &     $\dsc_{3}[32]$    \\   \hline\hline
                   %
%
 & \multicolumn{5}{c|||}{Batch Normalization}   &  & \multicolumn{5}{c||}{Batch Normalization}    \\\hline\hline
&  \multicolumn{5}{c|||}{$\poolmax_{5}[2]$} &  & \multicolumn{5}{c||}{$\poolmax_{5}[2]$}   \\\hline\hline              
     &   \multicolumn{5}{c|||}{ReLU}  & &   \multicolumn{5}{c||}{ReLU} \\\hline\hline
                                                 &    $\conv_{3}[32]$     &     $\conv_{3}[64]$     &     $\conv_{3}[48]$     &    $\conv_{3}[80]$     &     $\conv_{3}[64]$     &                              &    $\dsc_{3}[32]$     &     $\dsc_{3}[64]$     &     $\dsc_{3}[48]$     &    $\dsc_{3}[80]$     &     $\dsc_{3}[64]$       \\   \hline\hline
 & \multicolumn{5}{c|||}{Batch Normalization}   &  & \multicolumn{5}{c||}{Batch Normalization}    \\\hline\hline
  &  \multicolumn{5}{c|||}{$\poolmax_{5}[2]$} &  & \multicolumn{5}{c||}{$\poolmax_{5}[2]$}   \\\hline\hline
     &   \multicolumn{5}{c|||}{ReLU}  & &   \multicolumn{5}{c||}{ReLU} \\\hline\hline
                                                 &    $\conv_{3}[32]$     &     $\conv_{3}[32]$     &     $\conv_{3}[80]$     &    $\conv_{3}[48]$     &     $\conv_{3}[96]$     &                              &     $\dsc_{3}[32]$     &     $\dsc_{3}[32]$     &     $\dsc_{3}[80]$     &    $\dsc_{3}[48]$     &     $\dsc_{3}[96]$        \\   \hline\hline
 & \multicolumn{5}{c|||}{Batch Normalization}   &  & \multicolumn{5}{c||}{Batch Normalization}    \\\hline\hline
                   $\poolavg_{7}[\downarrow\! 3]$       &         \multicolumn{4}{g}{  $\poolavg_{5}[\downarrow\! 1]$}          &  $\poolavg_{5}[\downarrow\! 2]$       &                           $\poolavg_{7}[\downarrow\! 3]$       &         \multicolumn{4}{g}{  $\poolavg_{5}[\downarrow\! 1]$}          &  $\poolavg_{5}[\downarrow\! 2]$             \\   \hline\hline                   
                                      $\poolmax_{11}[\downarrow\! 5]$       &         \multicolumn{2}{c}{  $\poolmax_{5}[\downarrow\! 2]$}        &              \multicolumn{2}{g}{  $\poolmax_{7}[\downarrow\! 3]$ } &  $\poolmax_{5}[\downarrow\! 2]$   &    $\poolmax_{11}[\downarrow\! 5]$       &         \multicolumn{2}{c}{  $\poolmax_{5}[\downarrow\! 2]$}        &              \multicolumn{2}{g}{  $\poolmax_{7}[\downarrow\! 3]$ } &  $\poolmax_{5}[\downarrow\! 2]$
             \\   \hline\hline  
               \multicolumn{12}{||g||}{ 
               Densely Connected Neuron (DCN) matrix                
          $ \dense = \left(\dense_{k,\ell}\right)_{
1\leqslant k \leqslant K,
1 \leqslant \ell \leqslant \numparams 
 }$ 
multiplication operation for any $\netcrol_\bullet$. Outputs:
$
\netcrol_\bullet(\inputX)
=
\dense^{[\bullet]}
\times
\textrm{Vectorization} \left\{
\textrm{Input}
\right\}
$
               }  \\\hline\hline                 
                              \multicolumn{12}{||c||}{ $(a)$ Softmax operator denoted $\xi$ and yielding:}  \\
                              $\prob_{1}= \netcrol_{1}(\inputX)$       &     $\prob_{2}= \netcrol_{2}(\inputX)$    &     $\prob_{3}= \netcrol_{3}(\inputX)$    &     $\prob_{4}= \netcrol_{4}(\inputX)$    &     $\prob_{5}= \netcrol_{5}(\inputX)$    &     $\prob_{6= \netcrol_{6}(\inputX)}$    &     $\prob_{7}= \netcrol_{7}(\inputX)$    &     $\prob_{8}= \netcrol_{8}(\inputX)$    &     $\prob_{9}= \netcrol_{9}(\inputX)$    &     $\prob_{10}= \netcrol_{10}(\inputX)$    &     $\prob_{11}= \netcrol_{11}(\inputX)$    &     $\prob_{12}= \netcrol_{12}(\inputX)$       \\   \hline\hline
                                              \multicolumn{12}{||c||}{$(b)$ Softmax $\xi$ on categorical FCN with respect to Softlog $\newlog_\euler$, yielding:}  \\
                                \multicolumn{2}{||g|}{ $
\prob^{[13]}
 = 
\xi\left(
\bigoplus_{b=1}^2
\alpha_b
\newlog_\euler
\left(\prob^{[\bidx]}\right)
\right)$ }  &    \multicolumn{4}{c||}{  $
\prob^{[14]}
=
\netcrol_{147}(\inputX)
=
\xi\left(
\bigoplus_{b=3}^6
\alpha_b
\newlog_\euler
\left(\prob^{[\bidx]}\right)
\right)$  }   &           \multicolumn{2}{g|}{ $
\prob^{[15]}
=
\xi\left(
\bigoplus_{b=7}^8
\alpha_b
\newlog_\euler
\left(\prob^{[\bidx]}\right)
\right)$ }  &    \multicolumn{4}{c||}{  
$\prob^{[16]}
=
\netcrol_{16}(\inputX)
=
\xi\left(
\bigoplus_{b=9}^{12}
\alpha_b
\newlog_\euler
\left(\prob^{[\bidx]}\right)
\right)$  }       \\   \hline\hline\hline
                                                                                            \multicolumn{12}{||g||}{ Concatenation $[\prob_{1} || \prob_{2} || \cdots || \prob_{16}]$ from $(a)$ and $(b)$}  \\\hline\hline
                                                                                                                                          \multicolumn{12}{||c||}{ Softmax of categorical FCN with respect to Softlog $\newlog_\euler$, yielding:}  \\\hline
                                                                                                                                                                      \multicolumn{12}{||g||}{$\prob^{[17]}
=
\netcrol_{17}(\inputX)
=
\xi\left(
\bigoplus_{b=1}^{16}
\beta_b
\newlog_\euler
\left(\prob^{[\bidx]}\right)
\right)$}  \\\hline\hline                                                                                                   
\end{tabular}  
\end{scriptsize}
\label{tablesmodelsdragon0}
\end{table*}
 }
\end{landscape}%




 \section{
 Avoid using an explosive entropy formula
  }\label{sec softlog}


Let $\inputX(\omega)$ be an incoming feature associated with the input example $\omega$.
Let $\prob\oacc w \facc = \netcrol\left(\inputX(\omega)\right)$ be the output of network $\netcrol$ with respect to $\omega$.
The cross-entropy objective, derived from the so-called Shannon entropy, applies on small batches composed by several examples from:
\begin{equation}
\hspace*{-0.8cm}
{\cal E} \left( \prob \oacc w_s \facc  , \qrob \oacc w_s \facc \right) 
= 
- 
\frac{1}{S} \sum_{s=1}^S 
\sum_{k=1}^K \qrob_k  \oacc w_s \facc {\boldsymbol{\log}} \left( \prob_k \oacc w_s \facc  \right)
\label{eq entrop}
\end{equation}
where prediction states $ \prob\oacc w \facc$ are faced with true class distribution\footnote{
$\sum_{k=1}^K \qrob_k = 1$ and
$\qrob \in [0,1]^K$ in multilabel whereas 
$\qrob \in \{0,1\}^K$ in monolabel requirements.
}
 $ \qrob \oacc w \facc$, with $S$ denoting the batch size (number of training examples for a single iteration).

In classification, cross-entropy of Eq. \eqref{eq entrop} is used extensively as an objective, although experts in learning systems know that the use of softmax conditioning the derivation of $\prob$ is not in itself sufficient to prevent cross-entropy explosion.


\subsection{Problem: risk of cross-entropy explosion due to logarithm}

For very deep networks and/or a large number $K$ of output classes (the consequence is a large DCN matrix and thus some possible too-large output values for DCN), 
certain softmax values $\prob \oacc w_{s^\ast} \facc  = \xi \left( X (\omega_{s^\ast}) \right)$ associated with a particular example $w_{s^\ast}$ can, depending on the input data range or the initialization,  promptly reach a `0 in the numerical precision sense' and as a consequence, the cross-entropy contribution from a standard logarithm $\log \left( \prob \oacc w_{s^\ast} \facc  \right)$ on cross-entropy Eq. \eqref{eq entrop} is no more consistent with the computational updating stage involved in the learning process.

\subsection{ Literature findings }


Let $X$ be a DCN output as defined in Table \ref{tablesmodelsdragon0}.
Assume that set $X$ is sorted in a non-decreasing order. In this respect,
$X_{1} = \min \{  X_{k}  : k = 1, 2, \ldots, K \}$ 
and 
$X_{K}= \max \{  X_{k} : k = 1, 2, \ldots, K \}$.

In some deep learning models, 
ReLU[$X(\omega)$] is used instead of $X(\omega)$ for avoiding $X_{1} = \min \{  X_{k}  : k = 1, 2, \ldots, K \}$  to be a very large negative value. 
But this solution is questionable at this end-level stage dedicated to feature aggregation because a large number of features is set to the same value `0',  leading to less explainable frameworks in the sense that the minimum of softmax can be shared by a large number of examples pertaining to different classes (same class probability for several different categories,  somewhat far from reality).

Given that larger DCN matrices are more likely to deliver numerical issues (the maximum generated number has an expectation that of being larger as the size of the generated sample increases), another solution encountered in different models is to diminish the DCN size, that is reducing the second dimension $\numparams$ of the DCN matrix, a dimension relating the amount of input features to the DCN.
Indeed, a pooling operator can be applied for example to these input features to limit the number of exponential terms summed in the softmax denominator. 
But it is worth emphasizing that this solution is not relevant in terms of performance at large because $\numparams$ cannot be made smaller as desired without affecting performance: the larger the number of output classes,  the more complex the problem and the larger diversity of activations is required.

\subsection{Softlog solution proposed}

Why use an explosive entropy  when we could find alternative solutions based on customized logarithms ?
The rationale behind the customization proposed here is solely guided by numerical consistency.
We are interested in a parametric logarithm transform that has the form:
\begin{equation}
\newlog(x, {\alpha,\beta}) = 
\log 
\left( 
\alpha
x + \beta
\right)
\label{eq newlog}
\end{equation}
where scale $\alpha$ and shift $\beta$ parameters are selected, depending on the context, so as to embed the logarithm for a safe application with respect to numerical approximations on constrained input domains, randomized parameter initialization and recursive parameter updating strategies.
If we consider that the input measure set is a categorical probability interval $\measureset$ 
then, in contrast with yielding values in range $)-\infty, 0]$ as $\log$ does (the cause of numerical instability is the limit ``$-\infty$''), we consider output values in $[-1, 1]$. To this aim, and setting $\newlog(0, {\alpha,\beta}) =-1$ and $\newlog(1, {\alpha,\beta}) =1$ yields, with the notation $\euler=\exp{1}$, to:
$
\alpha =  \euler  - \euler^{-1}  
$
and
$
\beta = \euler^{-1}.
$
We just derived a \emph{softlog}: 
\begin{equation}
\boxed{
\newlog_\euler(x) = 
\log 
\left( 
\left(  \euler  - \euler^{-1} \right) 
x + \euler^{-1}
\right)
}
\label{eq logeuler}
\end{equation}
 that solves the main limitations caused by 
numerical approximations of $\log$(softmax($\bullet$)): 
initialization,  number of class outputs, minibatch size and the
rounding, numerical quantification, numerical conversion of data dynamics on $\{  X_{k}(\omega)  : k = 1, 2, \ldots, K \}$ 
have no of very less negative impacts on the consistency of Eq. \eqref{eq entrop} when $\boldsymbol{\log} \leftarrow \newlog_\euler$ in this equation.

\subsection{Softlog-softmax layers}\label{ssec integrators}

Softlog unlocks a door: the one that allows the use of multiple logarithms in the core of a learning system.
 Table \ref{tablesmodelsdragon0} shows several softlog integrators that operate directly on the softmax outputs $\prob^{[\bidx]}$. 
 Any community integrator receives softmax categorical probabilities, and applies successively the following operations: a softlog to distribute values in [-1, 1],  a weighted combination per class by using learnable weights,  a softmax operator.  
 This choice has been made in order to force the integrator to take into account the true decision states of the branch (otherwise, the inter-model comparison can be ineffective and, moreover, the combinations of weights can be inoperative thanks to variable data dynamics if the fusion is carried out at the level of the DCN weights).


It is important to notice here an additional role played by the softlog: without this operator,  we obtain 2 successive softmax (exponentiation of exponentially distributed data) and the cross entropy regularly diverges, while everything remains stable thanks to the insertion of the softlog between the multi-level multi-model softmax decision operators.

 \section{Do we need CAMs when we can obtain local decision states effortlessly ?}\label{sec sld}


We can, of course, be satisfied, as the ensemble learning literature is, to display a sub-selection of CAMs or the learned weights for knowing which feature/model contributes to the overall decision and to what extent. But this conceals a great deal of relevant information.

Consider individuals (sub-models) in a global community (ensemble model) challenge.
On the one hand, assuming that the community makes all its decisions by mimicking a specific individual that it identified as being the most relevant, then the problem is that the corresponding weight distribution (a Dirac function in this case) does not inform us about whether the other individuals learn consistently or not. The same issue holds true with CAMs.

Similarly, if the model results in a uniform distribution of weights (individuals' decisions are then deemed equivalent), then no information is measured on the fact that this is linked: 
\begin{itemize}
\item
either to individuals being equivalent on each sample, 
\item 
or not being equivalent per training sample, but to respective decisions being balanced on different proportions of the training samples where individuals are best per turn.
\end{itemize}

For these reasons, as well as a better understanding of the different contributions to the decisions, it is necessary to have a measure of the information provided by each individual to the community decision states. Such a tool can be a divergence measure: a variant deduced from the softlog operator is presented hereafter. The construction follows entropy and relative entropy designs, by highlighting specific properties of their softlog variants. We recall that $K$ denotes the number of categories.

\subsection{Softlog divergence}

Define the \underline{softlog entropy} by the operation: $E \mapsto - \newlog_\euler\left(  \prob(E)\right) $ where $E$ is an event and $ \prob(E)$ is the probability of this event.
The softlog entropy is respectively ``1 for an improbable event $E$'' and ``-1 for a certain event $E$'', while being monotonic with respect to $\prob(E)$, which is consistent with the definition of entropy as an uncertainty measure.
More specifically, the consistency of this definition simply follows from that $\mathcal{E}_{\euler}$ has the same monotony (increasing function) as the standard $\log$: the sole difference is that softlog entropy has finite (safe) bounds, in contrast with the standard entropy.

Let $Y$ denotes an individual or a community classifier, with
$Y \in \{
\netcrol_{1}, \netcrol_{2}, \ldots, \netcrol_{17}
\}$ for our purpose.
The \underline{softlog categorical entropy} $\mathcal{E}_{\euler}(Y)$ with respect to $Y$ is defined for an input example $x$ by:
\begin{IEEEeqnarray}{rCl}
\mathcal{E}_{\euler}\left(Y(x) \right)
&=&
- 
\sum_{k=1}^{K}
f_{Y}(x_k) \newlog_\euler\left(f_{Y}(x_k)\right) 
\label{eq entropylogeuler}
\end{IEEEeqnarray}
where
$f_{Y}(x_k) = \prob\left(Y(x) = k \right)$
is the probability that $Y$ associates example $x$ to category ``$k$''. 

In contrast with the Shannon entropy, we have the following property for the softlog entropy: 
\begin{equation}
\boxed{
-1 \leqslant 
 \mathcal{E}_{\euler}(Y)
\leqslant
1
}
\label{eq entp1}
\end{equation}

\begin{remark}
A Dirac distribution of probabilities on $Y$ yields
$
\mathcal{E}_{\euler}(Y)
=
-1 
$
whereas a uniform distribution implies
$
\mathcal{E}_{\euler}
=
- \newlog_\euler(1/K)
$
which is approximately ``1'' for very large values of $K$.
\end{remark}

The \underline{softlog categorical cross-entropy} of $Y^{[i]}$ with respect to $Y^{[j]}$ is defined by:
\begin{equation}
\mathcal{E}_{\euler}\left(Y^{[i]}(x) || Y^{[j]}(x) \right)
=-\sum_{k=1}^{K}
f_{Y^{[i]}}(x_k) \newlog_\euler\left(f_{Y^{[j]}}(x_k)\right) 
\label{eq crossentropylogeuler}
\end{equation}
We have furthermore: 
\begin{equation}
\boxed{
-1 \leqslant 
\mathcal{E}_{\euler}(Y^{[i]} || Y^{[j]} )
\leqslant
1
}
\label{eq crossentp1}
\end{equation}

\begin{remark}
For a Dirac distribution of probabilities on $Y^{[j]}$, noted as $f_{Y^{[j]}}(x_\ell) = \delta_{x_{k^\ast}}(x_\ell)$, we have:
$$
\mathcal{E}_{\euler}\left(Y^{[i]}(x) || Y^{[j]}(x)\right)
=
1 - 2f_{Y^{[i]}}(x_{k^\ast})
$$
thus if $Y^{[i]}(x)$ is also Dirac distributed $f_{Y^{[i]}}(x_\ell) = \delta_{x_{k^\dagger}}(x_\ell)$:
\begin{equation}
\mathcal{E}_{\euler}\left(Y^{[i]}(x) || Y^{[j]}(x) \right)
=
\left\{
\begin{array}{rcl}
1 & \textrm{~if~} &  k^\ast \neq k^\dagger \\
-1 & \textrm{~if~} & k^\ast = k^\dagger
\end{array}
\right.
\end{equation}
the latter case $k^\dagger=k^\ast$ meaning a perfect match between prediction and reality (class label).
\end{remark}

Finally, a \underline{relative softlog entropy} of $Y^{[i]}$ with respect to $Y^{[j]}$ is specified as a difference between cross-entropy and entropy, resulting in the expanded form:
\begin{IEEEeqnarray}{l}
\mathcal{D}_{\euler}\left(Y^{[i]}(x) || Y^{[j]} (x)  \right)
=
 \frac{1}{2}
\sum_{k=1}^{K}\!
 f_{Y^{[i]}}(x_k) \log \! \left( \!
 \frac{
 \left(  \euler^2  -  1 \right)  f_{Y^{[i]}}(x_k)  + 1
 }{
 \left(  \euler^2  -  1 \right) f_{Y^{[j]}}(x_k) +1
 } \!
 \right)
 \label{eq divKLlogeuler}
\end{IEEEeqnarray}
which does not admit any indetermination when a certain $f_{Y^{[j]}}(x_k)$ is approximately zero.  
The following symmetric version of this softlog relative entropy is called  \underline{SoftLog-Divergence} (SLD):
\begin{IEEEeqnarray}{rCl}
\label{eq divKLlogeulerSym}
\mathcal{D}_{\euler}\left(Y^{[i]} , Y^{[j]}   \right)
&=&
\frac{1}{2}
\left(
\mathcal{D}_{\euler}\left(Y^{[i]} || Y^{[j]}   \right)
+
\mathcal{D}_{\euler}\left(Y^{[j]} || Y^{[i]}  \right)
\right)
\end{IEEEeqnarray}

{SLD} $\mathcal{D}_{\euler}\left(Y^{[i]} , Y^{[j]}   \right)$ defined by Eq. \eqref{eq divKLlogeulerSym} verifies:
\begin{IEEEeqnarray}{l}
\boxed{
0 \leqslant \mathcal{D}_{\euler}\left(Y^{[i]} , Y^{[j]}   \right) \leqslant 1
}
 \label{eq divKLlogeulerbound}
\end{IEEEeqnarray}
This property in terms of normalized divergence is particularly important as it makes absolute or relative divergence comparisons possible (close or not to the min, close or not to the max, knowing that the maximum is a finite value unlike the Kullback-Leibler divergence).

\section{We should not hide the shortcomings of our models }\label{sec perfotensor}




In the early phase of deep learning, the literature focused on identifying the best CNN in terms of maximal achievable performance (for instance accuracy)  on a pre-specified dataset. This is largely due to the fact that there was a residual doubt about the ability of CNNs to solve intricate problems. 
Nowadays, the heuristic demonstration of our ability to find complex architectures for solving difficult problems is no longer an issue. Therefore, we can question this somewhat biased performance criterion: intrinsically, the learning performance is a stochastic process. Its stochasticity does not only depend on the variation of the dataset, it also depends on parameter initialization and hyperparameter setups for the optimization solver.

\subsection{
How to evaluate consistently in presence of randomized and manual conditioning ?
 }
 
Consistent comparison of stochastic processes requires comparing several statistics from the distributions of each pair of adjacent variables composing these processes. 
The trap is then to end up comparing a wide variety and a large number of statistics. To avoid this pitfall, we consider hereafter a reduced performance tensor to address the question of a template set relating the most salient performance fluctuations. 
We consider that any model, regardless of its complexity, must be characterized by such a performance template on initially simple problems, before eventually arguing that for complex problems, an excessively long training time does not justify numerous simulations.

 \subsection{Performance tensor solution}

Denote $Z$, the set of non-negative performance measures obtained when realizing a significant number of \emph{Monte Carlo} trials with respect to initialization randomness.
To provide a thorough analysis of the impact of parameterization on learning capabilities, we should provide, not only the maximum of $Z$ as usual, but also other informative statistics derived from the distribution of performance. 
The {performance tensor} discussed hereafter for describing the performance sample set $Z$ applies on
 $\minim, \meanim, \medium$ and $\maxim$ statistics associated with $Z$.
 It is written in \cite{atto21bookb} in the form:
\begin{equation}
\boxed{
\perfo(Z) =
\left[\myblue{\minim(Z)}, 
\left(\begin{array}{c} 
\meanim(Z)\\ 
\medium(Z)\\
\end{array} \right), 
\myred{\maxim(Z)}
\right]
}
\label{eq tensorperfo}
\end{equation}
A small $\minim(Z)$ is related to incapacity of learning depending on reasonable hyperparameters and/or initializations.
The centroid performance feature 
$\left(\meanim(Z), \medium(Z)\right)^\prime$
is such that:  $\meanim(Z) \gg \medium(Z)$ for situations where most learning variables have yield rather small performances.
 $\meanim(Z)$ can be very large even in this situation, but such episodic high performance on a model that lacks robustness is an undesirable property. 
A scalar performance metric that integrates with positive weights on the above tensor without the sole belief in max is the \emph{Ability}:
\begin{equation}
\boxed{
{\ability(Z) } =
\alpha
\maxim(Z)
+
\beta 
 \left[
\meanim(Z)
+
\medium(Z)
\right]/2
+
\gamma\:
\minim(Z)
}
\label{eq meantensorperfo}
\end{equation}
Thanks to such a combination, the max can play a higher role, but without forgetting the min observed  and, to a specified extent, the centroid.
We hope that the community will work out some dispassionate/consensual combination of $\{\alpha, \beta, \gamma\}$. 
In the meantime, we set $\alpha=1/2$, $\beta = \gamma = 1/4$.

  \subsection{ Virtues of the performance tensor illustrated on DRAGONFLY model}


From the performance tensor given in Table \ref{tab cnnresultsTestexpebdoqEpoch10iR001}, we can derive several virtues of the tensor approach on the DRAGONFLY model. 
On Experiment \#1, we observe the ability of DRAGONFLY to learn without any confusion after 10 epochs and whatever the 100 initializations performed on its parameters (min performance is 100\%), when the learning rates are $\{0.001, 0.005, 0.01\}$. 
If we had simply reported the maximum performance as usual, this behavior would have been hidden. 
Under the same max complacency, we would have reported 98\% on Experiment \#2 and 96\% on Experiment \#3: this would have hidden the fact that the min was often very low (78\% for Experiment \#3) and that the average performance can be up to 3\% away from the median (Experiment \#2 for example). 
More importantly, the inversion of the median position with respect to the mean when passing from Experiment \#2 (less difficult) to Experiment \#3 (more difficult) informs us that the intricacy of the problem can be a conditioning factor in determining the performance distribution shape/template.

{
\newcolumntype{g}{>{\columncolor[gray]{.95}[.5\tabcolsep]}c}  
\setlength{\arrayrulewidth}{0.35mm}
\begin{table}[!htp]
\centering
\caption{
DRAGONFLY performance tensor $ \perfo$ and ability $\ability$ in test for Experiments \#1, \#2, \#3.
} 
\begin{scriptsize}
\begin{tabular}{@{}c@{}c@{}c@{}}\hline\hline
\begin{tabular}{|| g | c  ||}
    \multicolumn{2}{c}{Experiment \#1 - Binary IO -  Epochs: 10 }  \\\hline
    $ \perfo$                                                                          &   $ \ability$        \\\hline\xrowht[()]{15pt}
     $\left[\myblue{100.0}, \left(\begin{array}{@{}c@{}} 100.0 \\ 100.0\\ \end{array} \right), \myred{100.0} \right]$    &    \textbf{100.0}
\\
\end{tabular}
&
\begin{tabular}{|| g | c  ||}
    \multicolumn{2}{c}{Experiment \#2 - Binary BDOQ -  Epochs: 10 }  \\\hline
        $ \perfo$                                                                          &   $ \ability $        \\\hline\xrowht[()]{15pt}
    $\left[\myblue{83.3}, \left(\begin{array}{@{}c@{}} 94.0 \\ 97.3\\ \end{array} \right), \myred{98.2} \right]$    &    \textbf{93.8}
\\
\end{tabular}
&
\begin{tabular}{|| g | c  ||}
    \multicolumn{2}{c}{Experiment \#3  - IO-SVA -  Epochs: 100 }  \\\hline
        $ \perfo$                                                                          &   $ \ability $        \\\hline\xrowht[()]{15pt}
    $\left[\myblue{77.7}, \left(\begin{array}{@{}c@{}} 86.9 \\ 84.5\\ \end{array} \right), \myred{95.7} \right]$    &    \textbf{88.7}
\\
\end{tabular}\\\hline\hline
\end{tabular}
\end{scriptsize}
 \label{tab cnnresultsTestexpebdoqEpoch10iR001}
\end{table}
 }





 \section{Divergence explained softlog-softmax ensemble learning}\label{sec expe}

This section addresses SLD relevance in terms of sub-community $\{\netcrol_{1}, \ldots, \netcrol_{16}\}$ learning, when training is performed ``ensemble'': all parameters are updated under $\netcrol_{17}$ objective variations.

We address evaluating individual/sub-community to the community decision $\netcrol^{[17]}$ from SLD
\begin{IEEEeqnarray}{rCl}
\label{eq divKLlogeulerSymSLDC}
\textrm{SLD}
\left(\prob^{[i]} , \prob^{[17]} \right)
=
\frac{1}{\numparams_{\Set}}
\sum_{x\in\Set}
\mathcal{D}_{\euler}\left(\prob^{[i]} (x), \prob^{[17]}  (x) \right)
\end{IEEEeqnarray}
where $\Set$ is a target set among training, validation, test data and ${\numparams_{\Set}}$ denotes the cardinal of set $\Set$.

Experimental results obtained with respect to 100 training epochs on the  IO-SVA Experiment \#3 (Section \ref{ssec geometrons}, see also Figure \ref{fig state1images}) are given in Table \ref{tablesmodels perfvs17}.

{
\newcolumntype{g}{>{\columncolor[gray]{.95}[.5\tabcolsep]}c}  
\setlength{\arrayrulewidth}{0.35mm}
\begin{table*}[!htp]
\centering
\caption{
Performance $\ability$ in \% and $\textrm{SLD}\left(\prob^{[i]} , \prob^{[17]} \right)$  for DRAGONFLY $\prob^{[17]}$ on Experiment \#3. 
} 
\begin{scriptsize}
\begin{tabular}{||     l          ||           g | c ||   g | c | g | c                      |||                  g | c ||    g | c | g | c                 ||}\hline\hline
                  \multicolumn{13}{c}{ Divergences (test) of community elements $\{\netcrol_{1}, \ldots, \netcrol_{16}\}$ with respect to $\netcrol_{17}$. $\ability\{ \netcrol_{17} \} = 89.7\%$. }  \\\hline\hline
  &\multicolumn{6}{g|||}{2D-C (standard 2D Convolutions)}  &  \multicolumn{6}{c||}{DSC (Depthwise Separable Convolutions)} \\\hline
& \multicolumn{2}{g||}{Generation \#1}  &  \multicolumn{4}{c|||}{Generation \#2}  &   \multicolumn{2}{g||}{Generation \#1}  &   \multicolumn{4}{c||}{Generation \#2}    \\\hline
People &$\netcrol_{1}$       &     $\netcrol_{2}$    &     $\netcrol_{3}$    &     $\netcrol_{4}$    &     $\netcrol_{5}$    &     $\netcrol_{6}$    &     $\netcrol_{7}$    &     $\netcrol_{8}$    &     $\netcrol_{9}$    &     $\netcrol_{10}$    &     $\netcrol_{11}$    &     $\netcrol_{12}$       \\   \hline
 %
$\ability$ &                   65.5       &    82.7    &     83.4    &     82.6     &    88.3    &     74.1     &     50.0      &   83.4     &   83.1    &    65.4     &    86.3     &    89.5    \\   \hline\hline                               
%
SLD       &           0.14       &   0.07    &    0.07     &   0.08     &    0.07    &    0.10     &     0.30      &   0.09     &   0.07    &    0.14     &    0.07     &     0.06    \\   \hline\hline                                                                                                  
Community &                \multicolumn{2}{g||}{ $\netcrol_{13}$  }  &    \multicolumn{4}{c|||}{  $\netcrol_{14}$   }   &      \multicolumn{2}{g||}{ $\netcrol_{15}$  }  &    \multicolumn{4}{c||}{  $\netcrol_{16}$   }      \\   \hline
 %
$\ability$  &                                   \multicolumn{2}{g||}{ 82.7  }  &    \multicolumn{4}{c|||}{  89.1   }   &      \multicolumn{2}{g||}{ 83.4  }  &    \multicolumn{4}{c||}{  89.4    }      \\   \hline
%
SLD       &                                \multicolumn{2}{g||}{ 0.08  }  &    \multicolumn{4}{c|||}{ 0.07   }   &      \multicolumn{2}{g||}{ 0.10  }  &    \multicolumn{4}{c||}{  0.06   }      \\   \hline\hline
%
%
%
                                              
\end{tabular}  
\end{scriptsize}
\label{tablesmodels perfvs17}
\end{table*}
 }

\begin{figure*}[!htp]
\centering
\includegraphics[width=0.7\textwidth]{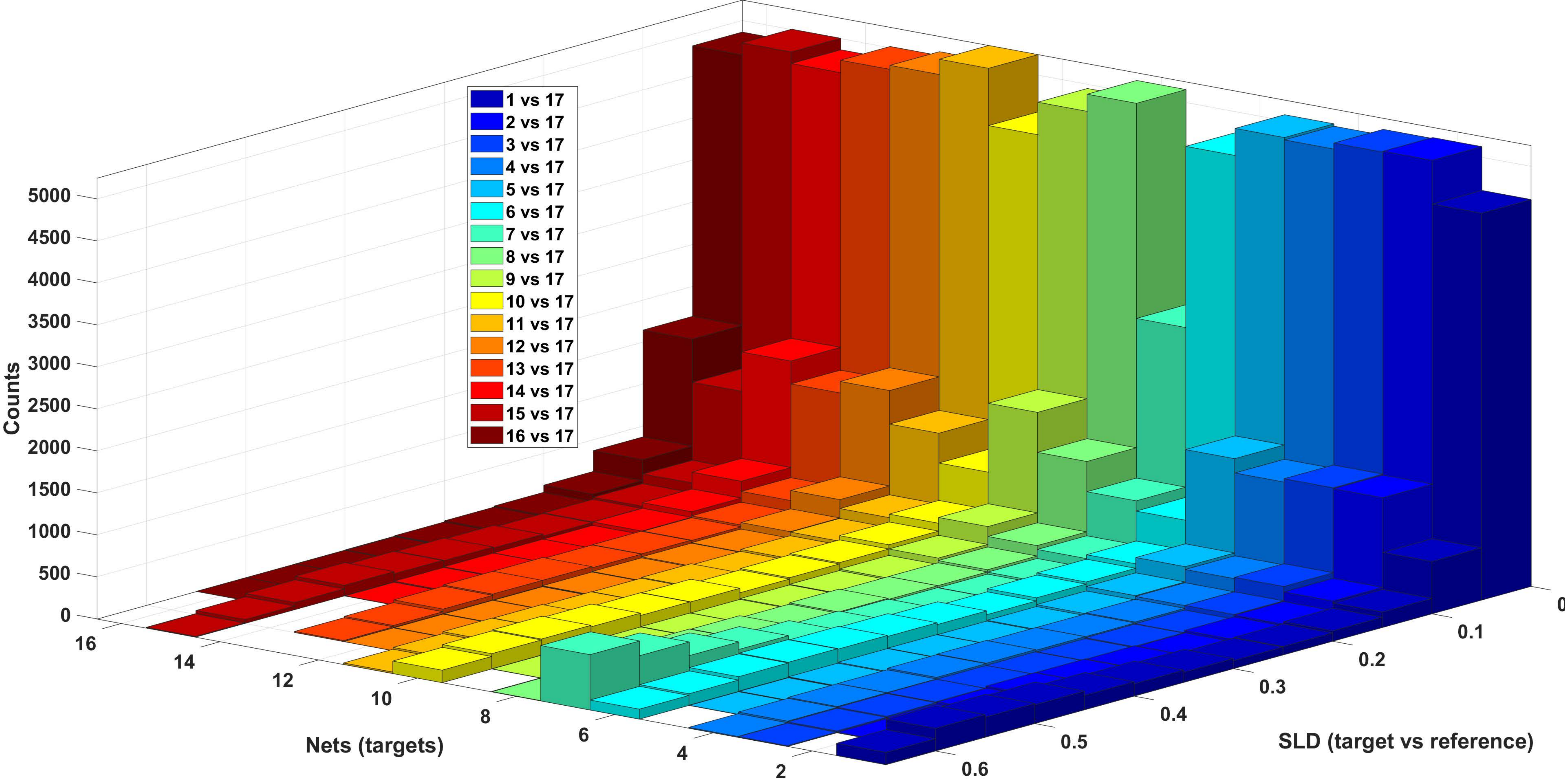}
\caption{
Histograms of 
$
\textrm{SLD}
\left(\prob^{[i]} , \prob^{[17]} \right)$ of Eq. \eqref{eq divKLlogeulerSymSLDC} 
 for Experiment \#3.
} 
\label{fig dragonkld}
\end{figure*}

Histograms of SLD are shown in Figure \ref{fig dragonkld}.
This SLD graphic is a powerful tool for analyzing contributions of $\netcrol_{\bullet}$ to $\netcrol_{17}$:
Large divergences are observed for
$\netcrol_{1}$ and $\netcrol_{7}$. Moderately large but numerous discordances are associated with  the trio: $\netcrol_{6}$,  $\netcrol_{10}$ and $\netcrol_{11}$. Very small divergences, except on rare examples, correspond to $\netcrol_{14}$ and $\netcrol_{16}$ (no divergence value larger than 0.5).

\begin{figure*}[!htbp]
\centering
\begin{tabular}{||@{~}c@{~}c@{~}|||@{~}c@{~}c@{~}||}\hline\hline
\begin{tabular}{@{}c@{~}c@{}}
\multicolumn{2}{@{}c@{}}{ $\prob(\inputX = \textrm{``O''})$}  \\ \hline\hline
\\
$\netcrol_{17}$  & 0.293 \\\hline
\\   $\netcrol_{6}$ & 1 \\ \\ $\netcrol_{9}$ & ${0.997}$ \\  \\ $\netcrol_{10}$ &  ${0.999}$  \\  \\  $\netcrol_{11}$  & ${0.997}$
\end{tabular}
&
\begin{tabular}{@{}c@{}}
\includegraphics[width=0.32\textwidth]{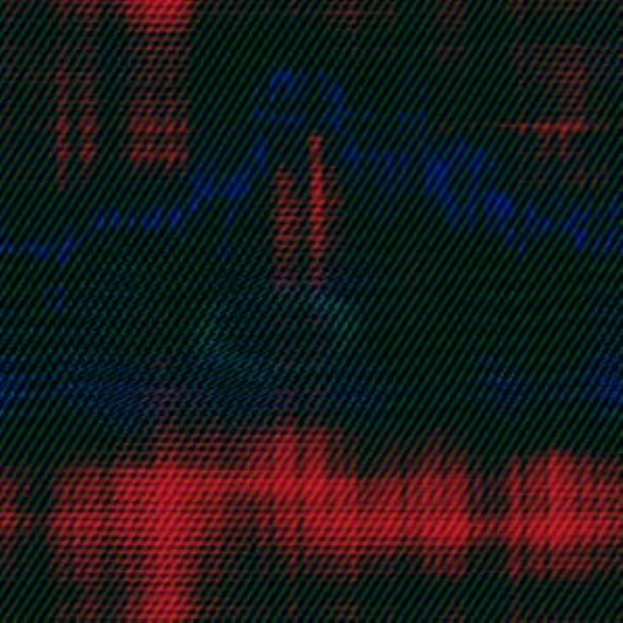}
\end{tabular}
&
\begin{tabular}{@{}c@{}}
\includegraphics[width=0.32\textwidth]{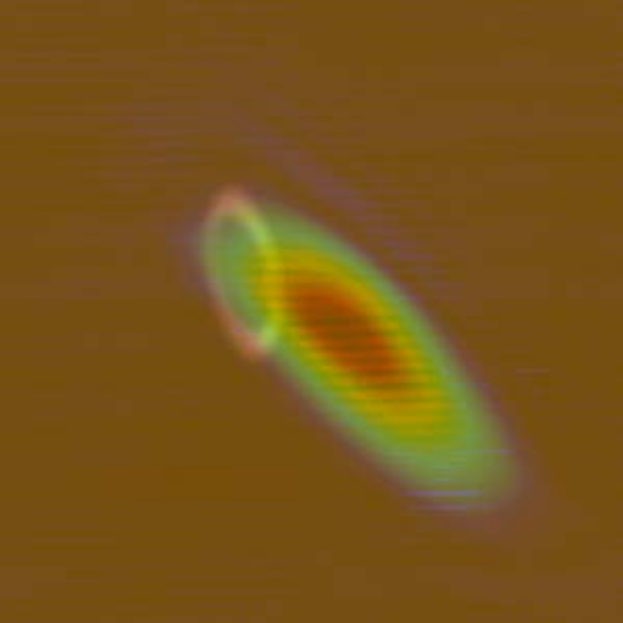}
\end{tabular}
&
\begin{tabular}{@{}c@{~}c@{}}
\multicolumn{2}{@{}c@{}}{ $\prob(\inputX = \textrm{``O''})$}  \\ \hline\hline
\\
$\netcrol_{17}$  & 0.463 \\\hline
 \\ $\netcrol_{4}$  & ${0.695}$ \\  \\ $\netcrol_{9}$ & ${0.864}$ \\ \\  $\netcrol_{12}$ & ${0.948}$  \\ \\ $\netcrol_{16}$ & ${0.998}$
\end{tabular} \\ \hline\hline
\end{tabular}
\caption{ 
Illustrative DRAGONFLY $\netcrol_{17}$ failures: it predicts ``I'' whereas the true label is ``O''. The majority of sub-models are wrong, excepted, per image, those highlighted at $(a)$ - $(b)$. 
} 
\label{fig dragonfails}
\end{figure*}

Finally, Figure \ref{fig dragonfails} shows situations where $\netcrol_{17}$ fails:
the first image $(a)$ is a discordance with the ``moderation'' trio $\{\netcrol_{6},  \netcrol_{10}, \netcrol_{11}\}$;   
 the second image $(b)$ is a discordance with the leaders $\{\netcrol_{12}, \netcrol_{16}\}$.



\bibliographystyle{IEEEtran}
\bibliography{BibFrustum}

\end{document}